\pdfoutput=1
\documentclass[12pt]{article}

\usepackage[letterpaper, margin=1in]{geometry}

\usepackage{amsmath, amssymb, amsthm}
\usepackage{graphicx}
\usepackage{booktabs}
\usepackage{multirow}
\usepackage{array}
\usepackage{tabularx}
\usepackage{longtable}
\usepackage{setspace}
\usepackage[colorlinks=true, linkcolor=blue, urlcolor=blue, citecolor=blue]{hyperref}
\usepackage{caption}
\usepackage{natbib}
\usepackage{titlesec}
\usepackage{tikz}
\usetikzlibrary{positioning, fit, calc, arrows.meta, backgrounds}

\titleformat{\section}{\normalfont\large\bfseries}{\thesection.}{1em}{}
\titleformat{\subsection}{\normalfont\normalsize\bfseries}{\thesubsection}{1em}{}

\newcommand{\MQ}{\mathrm{MQ}}

\newsavebox{\fitbox}
\newcommand{\ShrinkTabulars}{
  \let\origtabular\tabular \let\origendtabular\endtabular
  \renewenvironment{tabular}[2][c]{
    \begin{lrbox}{\fitbox}\origtabular[##1]{##2}
  }{
    \origendtabular\end{lrbox}
    \ifdim\wd\fitbox>\linewidth
      \resizebox{\linewidth}{!}{\usebox{\fitbox}}
    \else \usebox{\fitbox}\fi
  }
}

\title{\Large\textbf{Human Capital in Context:}\\
\textbf{What Employment Survival Reveals About Worker--Firm Match}\thanks{We thank
Stefan Elbl Droguett, Tianyu Du, Adam Keppler, Maureen McNichols, Valeri Nikolaev, Gerardo Perez Cavazos, Lawrence Schmidt, and Kayon Vafa for helpful
comments, and our research assistants, Khanh Vu and Quynh Mong. We also thank
the Data, Analytics, and Research Computing (DARC) team at Stanford GSB for technical
support. During the preparation of this work, we used Claude and ChatGPT to assist with
writing and data analysis; we reviewed and edited all content and take full
responsibility for it. All errors are our own.}}

\author{Jung Ho Choi\\
\textit{Stanford Graduate School of Business}
\and
Mario Milone\\
\textit{University of California, San Diego}
}

\date{November 2026}

\begin{document}

\hypersetup{pageanchor=false}
\begin{titlepage}
\maketitle
\thispagestyle{empty}

\begin{abstract}
\noindent There is no best worker in general, only a best worker for a
particular firm. We develop a measure of worker--firm match quality that treats
fit as a property of the \emph{pair}, learned from the survival of employment
relationships. A two-tower neural network embeds workers, from their career
histories and current positions, and firms, from their characteristics and
business descriptions, into a common space, and match quality is the cosine
similarity of the two vectors.
The measure takes its functional form from the skill-weights model of firm-specific
human capital, and its only estimated objective, a separation hazard, from the view of
the job match as an experience good.
Estimated on more than a billion person-periods covering 101 million workers and 8.4
million firms, a 0.1 increase in match quality lowers the odds of separation by 48\%.
About a quarter of the variance of match quality is specific to the
worker--firm pair, which worker-, firm-, and occupation-level measures
set to zero by construction. Workers earn more when better matched, especially early
in a relationship, and firms whose workforces score higher grow faster, become more
profitable, and are issued more, and more valuable, patents over the next five years
than their industry peers. Because it scores pairs that never
formed, the measure also predicts where movers go, and acquirers record a
larger share of the price as goodwill when the target's workers would fit the
acquirer better than they fit the target.


\end{abstract}
\end{titlepage}

\doublespacing

\setcounter{page}{1}
\hypersetup{pageanchor=true}

\section{Introduction}

Few decisions firms make are as consequential as the decision of who works
where. Hiring the wrong person, keeping the right person in the wrong role, or
losing a worker to a rival who can use their skills better dissipates value at
least as surely as a bad capital budgeting decision, and more surely still in the firm
\citet{zingales_search_2000} describes, whose critical asset is its human capital
rather than its physical assets. The stakes do not
stop at the firm's boundary. Aggregated across
firms, these decisions determine where the economy's human capital is put to work
\citep{abowd_relation_2005}. A long literature accordingly
establishes that management quality and human resource practices are
first-order determinants of productivity and firm value
\citep{bloom_human_2011, edmans_does_2011}. Yet the resource
that managers most often describe as their most valuable asset enters most
empirical work as a scalar (headcount, wage bill, or intangible-capital
proxy), when it enters at all. Nothing in standard reporting, and little in
existing measurement, reveals whether the people inside a firm are the right
people \emph{for that firm} \citep{honigsberg_wage_2022}.

The phrase ``for that firm'' carries the substance of this paper. In much
empirical work, a good hire is simply a good worker: more educated, more
experienced, more able. But the idea that the value of human capital depends
on where it is deployed is among the oldest in labor economics
\citep{becker_human_1994}, and its modern statement, the skill-weights model
of \citet{lazear_firmspecific_2009}, turns it into a precise claim about
matching: skills are multidimensional and general, but each firm weights the
dimensions idiosyncratically, according to its predetermined organizational
structure, its culture, and its business model. On this view there is no best
worker in general, but only a best worker for a particular firm. The same
engineer is a different asset to a research-intensive firm than to a general
management consultancy, and two firms competing for that engineer are not, in general,
equally good homes for those skills. What matters for hiring, for retention, and
for the market for talent is therefore a property of the worker--firm
\emph{pair}, and a measure that attaches quality to the worker alone---or to
the job alone---cannot capture it.

That object has gone unmeasured for reasons that are practical rather than
conceptual. The empirical literature on human resource management grew up
around what could be observed: incentive structures, monitoring, and targets
can be elicited from managers through surveys, while the quality of
assignment---who is matched to what---cannot, so the assignment margin
received far less scrutiny than its economic importance warrants.
\citet{coraggio_jaq_2025} broke this constraint by combining administrative
employer--employee data with machine learning, measuring whether each worker
holds the occupation their characteristics best suit and showing that the
resulting allocation quality correlates with productivity and improves with
managerial quality. Their measure establishes that allocation is measurable
and economically meaningful, but it is defined over \emph{occupations} and is
therefore invariant to the employer: it can say that a worker is in the wrong
job, not that the worker is at the wrong firm. A second constraint was how workers
and firms are described. Economists have traditionally summarized firms by a few
observable characteristics, such as industry, size, or accounting ratios, and workers
by education and experience. A handful of categories cannot carry a firm-specific
weighting of many skill dimensions. Machine learning has relaxed this constraint, but no existing work describes both
sides of the employment relationship in the same terms, so that a worker and a firm
can be compared directly, which is the prerequisite for measuring the pair.

We introduce the \textit{Human Capital Transformer} (HCT), which describes both sides
in the same terms. A transformer, the model
architecture behind modern language models, reads a worker's pre-match career history
as a sequence of positions, much as a language model reads a sentence word by word. Its
summary of that history, combined with static characteristics and the worker's evolving
position inside the current firm, is mapped into an \emph{embedding} (also called a
representation): a vector of numbers that summarizes the worker and places similar
workers close together. A second model builds
the firm's embedding from its characteristics and the text describing its business,
producing a vector of the same length. The two models are called towers, and the
architecture a two-tower network, because each side is processed separately until the
two vectors meet. Match quality, $\MQ_{wft}$, is the cosine similarity between the two
vectors (a weighted cosine, Section~\ref{sec:mq}): the cosine of the angle between
them, which is high when the worker's profile points in the direction the firm
rewards. Under the normalization of Section~\ref{sec:multidim},
it is exactly the inner product $\omega_f^{\top} s_w$ that the skill-weights model of
\citet{lazear_firmspecific_2009} implies. Two
properties follow from this construction. Because match quality is a property of
the pair rather than of the worker alone, the same worker can be a strong match
for one employer and a weak match for another. Because both towers are
functions of characteristics rather than identities attached to a
relationship, the score is defined for any worker--firm pair, including pairs
that are never observed: it prices counterfactual reallocations---what a target's
workforce would be worth to a particular acquirer, or which of two competing
offers is the better home for a given engineer.

The identifying idea is that good matches reveal themselves through their
\emph{durability}. Following \citet{jovanovic_job_1979}, the employment
relationship is an experience good: its quality is learned as it is lived, and
poor matches dissolve while good matches endure. We therefore treat each
worker--firm relationship as the unit of observation, one row (a \emph{person-period})
per year of tenure, and embed $\MQ_{wft}$ in a discrete-time separation hazard,
\begin{equation*}
\lambda_{wft} = \sigma\!\big(b(t) + \beta\,\MQ_{wft}\big), \qquad \beta < 0,
\end{equation*}
where $b(t)$ is a flexible baseline profile in tenure and $\sigma$ is the
logistic function. The tower parameters, which determine $\MQ_{wft}$, and $\beta$ are
estimated by maximizing the resulting likelihood, with the baseline fit alongside them.
Two-tower networks are usually trained differently: in a recommendation system, the
model learns to rank the item a user actually chose above items the user did not, and
the analogue here would be to predict a worker's next employer. We depart from that
objective for a reason specific to employment data: most consecutive observations of a
worker involve no change of employer, so what the data mostly record is persistence,
which is better modeled as duration than as a correct guess about the next employer
(Section~\ref{sec:unit}). Estimation uses \emph{only}
the timing of separations. No wage, no productivity measure, no destination
employer, and no firm financial variable enters the objective, so these outcomes
remain available for validation, and the circularity that a benchmark-based
construction must argue against does not arise.\footnote{Duration is an imperfect
proxy for quality in two opposing directions. A poor match may persist because moving
is costly, and an excellent match may end because it succeeded.
Section~\ref{sec:duration-limits} explains how we confront both.} In sum, the HCT
departs from a standard two-tower recommendation model in three ways: the worker is
read as a career sequence rather than a static profile; the cosine between the two
embeddings is not a modeling convenience but the inner product the skill-weights model
implies, which fixes the form of the measure and the scale of $\beta$; and the
objective is the survival of the relationship rather than the retrieval of the next
item.

The empirical work proceeds in three sets of tests, which together cover the five
predictions of Section~\ref{sec:strategy}. We first ask whether
the measure predicts separations, the outcome it is built on, and whether it
is specific to the pair and not only to the worker or the employer.
We then examine whether the pair-level object is priced and acted upon
outside that setting, in wages and in where movers go. We finally aggregate the measure to
the firm, and relate it to performance, to innovation, and to the price paid
for an acquired workforce.

Estimated on the full Revelio Labs (hereafter Revelio) universe (1.18 billion
person-period observations covering 101 million workers and 8.4 million
firms), the hazard coefficient is $\hat\beta = -6.49$: a 0.1 increase in match
quality lowers the odds that a relationship ends in a given year by 48\%. The
coefficient is not a coefficient on the business cycle or on observable
composition: entering the trained score in a logit that adds calendar-year
fixed effects to the tenure baseline, and then observable worker and firm
controls, barely moves it. Nor does the score merely fit the workers it was trained
on. On held-out workers, whom the model has never seen, we measure predictive accuracy
by the AUC. Adding $\MQ$ raises the AUC for separation from 0.695 for the tenure
baseline to 0.764. Relative to a logit that already conditions on tenure, calendar year,
and observables, it raises the AUC from 0.734 to 0.778, above the HCT's own
0.764, so the HCT's entire advantage survives as a single linear term in a conventional
hazard.

Within the first set of tests, the paper's central result is what the score is made of.
Regressing $\MQ$ on worker fixed effects explains 50\% of its variance,
firm fixed effects 44\%, and the two
entered additively 73\%. The remaining 27\% (25\% on held-out workers) is
pair-specific: it cannot be
attributed to some workers being generally well matched or to some firms
generally matching their workers well, and the residual has a standard
deviation about half that of the score itself. This
is the restriction the skill-weights model implies and that worker-level,
firm-level, and occupation-level measures set to zero by construction.

The second set of tests asks whether this pair-level object is priced and
acted upon in decisions the model never saw. Wages are the price at which a
pairing is transacted, and a measure of fit should be reflected in them:
workers earn about 7.9\% more per 0.1 of match quality, 3.4\% conditional on
observables, and 1.3\% under worker and firm fixed effects jointly---the
specification in which only the pair component of both wages and $\MQ$
survives. At every level of control the association is larger in the first years
of a relationship than later on, which is where matching theory says the surplus is
being revealed and priced. The match-specific premium fades to zero at long tenure,
consistent with durable relationships also reflecting the
cost of moving.
Destination
choice is the harder test, because it asks the score to order pairs that did
not exist when it was estimated. Scoring each mover against a thousand
time-valid candidate employers, the model places the realized destination
first 6.8 times and in its top ten 3.4 times as often as chance, despite never
having been trained to retrieve employers at all. Neither result follows from
the separation fit mechanically. A score that merely tracked inertia would
predict who stays, but neither what a worker is paid nor which of a thousand firms
the worker joins next.

The third set of tests asks whether the measure aggregates to the firm. It
does.\footnote{Because the model is estimated on the full sample period, later
information could leak into earlier scores. We therefore also estimate the model with a
split in time rather than across workers, training on data through 2016 only, and
relate match quality measured in 2016 to outcomes from 2017 onward (Appendix
Tables~\ref{tab:firmhor2016}, \ref{tab:innovhor2016}, and~\ref{tab:ma2016}). The results
are qualitatively the same.}

Within industry and year, a 0.1 increase in firm match quality is
associated with 5.4 percentage points faster revenue growth and 4.2 points
faster employment growth over the next year, and with a 1.4 point larger rise
in return on assets, holding fixed size, R\&D, market value, book-to-market,
and leverage---so the result is not that well-matched firms are highly valued
firms. Within firm over time the growth associations remain large, and the
profitability association appears from the second year on. Because
innovation arrives with a lag, we relate the same firm-level score to the patents issued to a
firm over the next five years and to the citations and value of each patent.
Within industry and year, better-matched workforces are issued more patents, and
their patents are more valuable and more cited; the value association
holds within firm as well, the count association holds within firm only at the
five-year horizon, and within firm the citation association appears only for the
earliest patents.

Human capital is part of what an acquirer buys \citep{lee_human_2018,
chen_human_2021, kim_startup_2024, lagaras_mas_2026}, and the last test in the third
set asks whether the match quality measure can capture it. When one firm acquires another, the value of
the acquired workforce is folded into goodwill, an unmeasured residual by
construction. We measure the goodwill and purchase price that acquirers report for
each acquisition in eXtensible Business Reporting Language (XBRL) filings. The test
exploits a property that no match fixed effect has. Because $\MQ$ is a function of
worker and firm inputs rather than of the relationship, we can evaluate a
target's workers not only against the target but against the acquirer they are
about to join. Across 1,334 acquisitions with at least 20 scored target relationships,
the better the target's workers
would match the acquirer relative to how well they matched the target, the larger the
share of the purchase price the acquirer records as goodwill, conditional on
acquirer match quality, acquirer financials, and acquirer-industry,
target-industry, and year fixed effects. What the acquirer appears to pay for
is the pairing the deal creates relative to the one it dissolves.

The contribution has three parts. The first, and the most important, is the finding:
the object measured is the pair, and the pair component is first-order. The
wage-decomposition tradition since \citet{abowd_high_1999} treats log wages as
additive in worker and firm effects and typically finds the match-specific interaction
to be small; the mismatch literature measures fit as the distance between a worker's
skills and the requirements of an occupation or the profile of coworkers
\citep{fredriksson_mismatch_2018, guvenen_multidimensional_2020,
lise_multidimensional_2020}; and the closest antecedent, the allocation-quality measure
of \citet{coraggio_jaq_2025}, is defined over occupations. We identify the match
component from durability rather than from wages and find that it accounts for about a
quarter of the variance in fit and shows up in retention, wages, mobility, firm
performance, and the prices paid for workforces. The same worker is a materially better
fit for some employers than for others, and that difference is priced and acted upon.
The second is that the measure is usable well beyond the relationships it was estimated
on. Because both towers are functions of characteristics rather than identities, the
score is computable for any worker and any firm, whether or not they have ever been
observed together. It aggregates from the bottom up: a firm-level average of pair
scores, built from separation timing alone, is associated with growth, profitability,
and innovation. It evaluates counterfactual pairings, such as a target's workforce at
its acquirer, which no fixed effect attached to a realized relationship can do; for
accounting and corporate finance, that puts a number on an asset that financial
reporting deliberately leaves inside goodwill, the assembled workforce. And the worker
and firm embeddings themselves are available as inputs to other studies. The third is
the construction. The measure is theoretically grounded rather than ad hoc: its
functional form is the inner product of the skill-weights model of
\citet{lazear_firmspecific_2009}, and its objective is the separation hazard that the
experience-good view of the job match \citep{jovanovic_job_1979} implies; the model is
left free to learn the content of the skill dimensions, not the structure of the
measure. The estimation combines a transformer over career sequences, in the spirit of
the CAREER model of \citet{vafa_career_2024} and its language-model successor
\citep{athey_laborllm_2024}, with firm embeddings learned from text and
characteristics, in the spirit of the asset embeddings of \citet{gabaix_asset_2025},
inside a two-tower network whose recommendation-style objective is replaced by a
survival likelihood. To our knowledge, no prior work combines these three elements, a
sequential encoder for one side of the pair, a similarity fixed by theory, and a
survival objective, and the combination is what makes the score meaningful as a measure
rather than as a predictor.

The paper proceeds as follows. Section~\ref{sec:lit} reviews three related
literatures: human capital measurement; job matching, turnover, and allocation quality;
and machine learning. Section~\ref{sec:theory} develops the theoretical foundation of
the measure, combining multidimensional skill matching with the view of the job match
as an experience good that motivates the hazard, discusses the limits of duration as a
signal, and maps the framework into embeddings.
Section~\ref{sec:hct} describes the architecture and the estimated objective.
Section~\ref{sec:data} presents the data, the research design, and the
empirical predictions and validation strategy that organize the results.
Section~\ref{sec:results} reports the results in the three sets described
above: (i) hazard estimates, predictive performance, and the pair decomposition;
(ii) wages and destination choice; and (iii) firm performance, innovation, and
acquisitions. Section~\ref{sec:conclusion} concludes with the interpretation
and limitations of the measure and the uses to which the embeddings can be put.

\section{Literature Review}\label{sec:lit}

This section positions our work within three literatures: human capital measurement;
job matching, turnover, and allocation quality; and machine learning. The first two
supply the object and the identifying idea, while the third supplies the tools and shows what
has not yet been done with them.

\subsection{Human Capital Measurement}

The foundational work on human capital measurement begins with
\citet{becker_human_1994}, who developed the economic theory of human capital
investment, and \citet{mincer_schooling_1974}, whose earnings-experience equation
became the workhorse model in labor economics. These studies established education
and experience as the primary observable measures of human capital, enabling decades
of research on the returns to schooling and on-the-job training. They remain
the field's measures: recent surveys of what economists have learned about human
capital \citep{deming_four_2022} and of how it is measured
\citep{abraham_measuring_2022} find that, beyond schooling, experience, and
earnings-based aggregates, few operational measures exist, and none that attaches
value to a worker in a particular place of work. The closest attempt to build a
firm-level measure from workers is \citet{abowd_relation_2005}, who use matched
employer--employee records to construct the human capital of each firm's workforce
from person effects estimated in wage equations and relate it to productivity and
market value; the measure is additive in worker quality by construction. A growing
body of work questions whether scalar measures of this kind capture the
multidimensionality of human capital. Skills are many, technology and trade shift
the returns to them differentially \citep{autor_skill_2003, deming_growing_2017},
and, more importantly, firms differ in which skills they demand:
\citet{deming_skill_2018} document large variation in skill requirements across
firms within the same occupation and labor market, and \citet{hershbein_recessions_2018}
show that the requirements themselves move with the business cycle.
\citet{lazear_firmspecific_2009} provides the theory that organizes these observations. In his
skill-weights framework, firms place heterogeneous weights on general skill
dimensions, so that the productivity of a given bundle of skills, and hence the
firm-specificity of human capital, is a property of the worker--firm pair.

If firms weight skills differently, then the same human capital is valued
differently in different places, and a body of work confirms it.
\citet{sattinger_assignment_1993} provides the theoretical framework for skill
distributions and assignment patterns, emphasizing that workers with different
skill profiles sort into different occupations and firms.
\citet{gibbons_taskspecific_2004} develop a model of task-specific human capital,
in which skills acquired in one context transfer imperfectly to others, so that
career trajectories carry information about accumulated capabilities.
\citet{gathmann_how_2010} measure task-specific skills from worker--task assignment
data and document substantial heterogeneity in skill transferability across
occupational transitions. These studies establish that the value of a skill bundle
depends on the context in which it is deployed, but they measure that dependency
at the level of tasks and occupations and infer it from wage changes at
transitions. They therefore point to an object they do not measure: how well a
particular worker fits a particular firm, evaluated before or after the fact,
for pairs that formed and pairs that did not.

The matched employer--employee literature comes closest to this object and,
instructively, treats it as a residual. \citet{abowd_high_1999} decompose earnings
into additive worker and firm effects, revealing substantial heterogeneity in firm
wage premiums. \citet{card_workplace_2013} extend the approach to rising wage
inequality in West Germany, attributing part of it to assortative matching between
high-skill workers and high-wage firms, and \citet{song_firming_2019} document the
concentration of highly paid workers in superstar firms. In each case the
worker--firm interaction is what is left after additive effects are removed, and
it is typically found to be small in wages. The decompositions are also known to
overstate the additive components when few workers move between firms, the limited
mobility bias of \citet{andrews_high_2008} that \citet{kline_leaveout_2020} and
\citet{bonhomme_how_2023} correct; Section~\ref{sec:pair} explains why a measure that
does not rely on movers to identify firm effects is less exposed to it. On the accounting side,
\citet{demers_measuring_2024} and \citet{batish_human_2021} measure what firms
disclose about their human capital, and \citet{honigsberg_wage_2022} show how little
current accounting rules require firms to report about their workforces. Disclosure
describes what firms say about their workforces rather than what those workforces are
worth to them. None of these literatures produces a measure of the pair.
Section~\ref{sec:theory} takes the skill-weights model as the definition of that
object, and Section~\ref{sec:hct} builds the measure.

\subsection{Job Matching, Turnover, and Allocation Quality}

Our identification strategy draws on the tradition that treats the employment
relationship as an experience good whose quality is revealed over time.
\citet{jovanovic_job_1979} models job matching as a process of learning about an
unobserved match-specific component: workers and firms accumulate information about
the quality of the pairing, poor matches dissolve, and good matches survive. The
central empirical implication---that separation hazards decline with tenure because
surviving matches are selected on quality---is the foundation for our behavioral
model. \citet{taber_estimation_2020} develop estimation approaches for models of
this class.

A more recent literature measures mismatch directly. \citet{guvenen_multidimensional_2020}
and \citet{lise_multidimensional_2020} define it as the distance between a worker's
multidimensional skills and the requirements of the occupation, and show that
mismatch lowers wages, slows human capital accumulation, and is gradually
corrected through mobility. \citet{fredriksson_mismatch_2018} define it as the
distance between a worker's cognitive and non-cognitive skill profile and that of
their coworkers, and find that it raises separations, most strongly for
inexperienced workers about whom least is known at hiring, and is priced into
entry wages only for workers with a track record. These papers establish two
facts on which we build: mismatch is real and costly, and separations reveal it.
Their measures, however, require test-based skill data that exist only for
particular cohorts, and they define the requirement side at the level of the
occupation or the coworker group rather than the firm.
The identifying idea we take from them---that the hazard of separation is
informative about fit---can be applied without skill tests, to everyone, and at
the level of the firm, which is what the survival objective of
Section~\ref{sec:objective} does.

Directly relevant is the recent work that measures allocation quality
empirically. \citet{coraggio_jaq_2025} construct the job allocation quality (JAQ) measure using
machine learning methods on Swedish register data, benchmarking worker--occupation
assignment against the allocation observed in the most productive firms, and show
that JAQ correlates with productivity, competition, ownership structure, and
managerial quality. Our approach differs in the object measured (worker--firm fit
rather than worker--occupation suitability), the identifying variation (relationship
survival rather than a productivity-selected benchmark), and the data requirement
(public career histories rather than administrative registers).

\subsection{Machine Learning}\label{sec:replearning}

Machine learning methods have been applied with particular success to prediction
and measurement.
\citet{erel_selecting_2021} use machine learning to predict board director
performance and show that algorithmic selection improves on the choices firms
actually make. In accounting, \citet{bao_detecting_2020} predict fraud from raw
accounting numbers with ensemble methods, \citet{chen_predicting_2022} predict
the direction of earnings changes from detailed XBRL financial data and show that
the predictions earn abnormal returns, and \citet{geertsema_relative_2023} use
gradient-boosted trees for relative valuation and peer selection. A second use is
measurement rather than prediction: \citet{kogan_technological_2017} construct
market-based measures of the value of patents, which we use as an outcome, and
\citet{demers_measuring_2024} construct measures of human capital disclosure from
text. What this literature has in common is that the objects it predicts or
measures belong to the firm or the security. What it has not done is bring the
same tools to the labor side of the firm. No study has represented workers and firms
in a common space so that their fit can be measured; the closest,
\citet{vafa_career_2024} and \citet{athey_laborllm_2024}, discussed below, represent
workers alone. Representation learning, the branch of machine learning that learns
such representations from data, supplies the tools for that step.

Representation learning has changed how researchers extract patterns from
unstructured data. \citet{mikolov_distributed_2013} introduce word embeddings,
demonstrating that distributional patterns encode semantic information.
\citet{vaswani_attention_2017} introduce the transformer architecture,
\citet{devlin_bert_2019} introduce BERT, and \citet{reimers_sentencebert_2019}
extend it to sentence-level embeddings, which we use to encode firm descriptions.
\Citet{oord_representation_2019}, \citet{he_momentum_2020}, and
\citet{chen_simple_2020} develop contrastive representation learning.
\citet{vafa_career_2024} apply a transformer encoder to job title sequences to
produce career embeddings, showing that a foundation model of labor sequence data
predicts occupational transitions better than conventional models;
\citet{athey_laborllm_2024} extend the approach by fine-tuning a large language model on
career histories represented as text; and \citet{gabaix_asset_2025} apply representation learning to assets, recovering firm
representations from investors' portfolio holdings.

Two-tower architectures embed two types of entities into a shared latent space,
with relevance measured by inner product or cosine similarity
\citep{yang_mixed_2020}. In their standard application to recommender systems, these
architectures are trained with contrastive objectives that score observed pairs
against sampled alternatives, because, in that setting, a user's continued
engagement with an item is not observed and negatives must be constructed. Our use
of the architecture retains the two towers and the cosine similarity between them but
replaces the retrieval objective with a survival model, for reasons
Section~\ref{sec:unit} develops, and treats the score's performance on next-employer
retrieval as evidence rather than as something to be optimized.

\section{Theoretical Foundation of the Match-Quality Measure}\label{sec:theory}

This section develops the framework behind the measure. It has two
components: a static account of what match quality \emph{is}
(Section~\ref{sec:multidim}) and a dynamic account of how it becomes
\emph{observable} (Section~\ref{sec:revelation}). The static account disciplines the
\emph{functional form} of the measure; the dynamic account disciplines the
\emph{objective function} used to estimate it. Both are needed: a theory of what a
good match \emph{is} places no restriction on which moment of the data should recover
it, and a theory of \emph{when relationships end} places no restriction on the form of
the score.

\subsection{Multidimensional Skill Matching}\label{sec:multidim}

Following \citet{lazear_firmspecific_2009}, we model firms as placing heterogeneous
weights on multidimensional human capital. Let worker $w$ have a $K$-dimensional
skill vector $s_w = (s_{w1}, \ldots, s_{wK})$, where each dimension captures a
distinct aspect of human capital such as technical expertise, managerial ability, or
industry-specific knowledge. Let firm $f$ have a task-weight vector
$\omega_f = (\omega_{f1}, \ldots, \omega_{fK})$ reflecting the relative importance of
each skill dimension to its production technology.\footnote{We normalize
$\|\omega_f\|_2 = \|s_w\|_2 = 1$, so that $\omega_f$ captures the composition of a
firm's skill demands rather than its scale; without this restriction the level of
$y_{wf}$ would absorb firm productivity, and its scale would not be separately
identified from the hazard coefficient $\beta$ introduced in
Section~\ref{sec:revelation}. \citet{lazear_firmspecific_2009} instead normalizes the
weights to sum to one, which reads naturally as skill shares; we use the Euclidean
normalization because it bounds match quality in $[-1,1]$ and thereby fixes the scale
of $\beta$.} Firm-specific productivity for
worker $w$ at firm $f$ is the inner product
\begin{equation}
y_{wf} = \omega_f^{\top} s_w .
\end{equation}
This formulation embeds the \citet{roy_thoughts_1951} model of comparative advantage
as a special case: workers sort into firms where their particular combination of
skills is most valued. That the advantage is comparative rather than absolute is what
matters for measurement: were skill one-dimensional, every worker would rank firms
identically, sorting would be purely vertical, and match quality would collapse into
additive worker and firm effects.

The key implication for measurement is that match quality is inherently a function
of both worker skills and firm requirements. Two workers may share a job title and
differ in the composition of their skills, one stronger in statistics and the other in
software engineering. A firm that runs clinical trials values the first more highly,
while a firm whose product is a consumer application values the second, so that the two
firms rank the same pair of workers in opposite order. This interaction makes it insufficient
to measure human capital as a scalar attached to the worker, and it is the reason a
measure defined over occupations alone---invariant to the employer---cannot capture
the object of interest. It also delivers a testable restriction: if match quality is
genuinely a property of the pair, a non-trivial share of its variance must survive
the removal of additive worker and firm effects.

\subsection{Duration as the Revealing Moment}\label{sec:revelation}

The static account fixes what match quality is but leaves it unobserved, in two
distinct senses. To the econometrician, $\omega_f^{\top} s_w$ is latent: neither the
skill vector $s_w$ nor the task weights $\omega_f$ appears in any dataset, so the
measure must be recovered from some observable moment. To the parties themselves,
match quality is imperfectly known at the moment of hiring: a resume and an interview
are noisy signals of how a particular worker will perform inside a particular
organization. Following \citet{jovanovic_job_1979}, we treat the employment
relationship as an experience good in this second sense: its quality is learned as it
is lived, and the parties act on what they learn.

The two senses do different work. The first makes an observable moment necessary, while
the second makes duration the informative signal. Were match quality known at hiring and
matching frictionless, only good pairings would form, realized $\MQ$ would be truncated
from below, and separations would trace shocks rather than quality. It is precisely
because the parties must learn about match quality that poorly matched pairs form, are
discovered, and dissolve,
generating the dispersion in $\MQ$ across observed relationships that the hazard
exploits. A worker $w$ remains with firm $f$ so long as the value of the relationship
exceeds the value of the outside option net of mobility costs, and separates when it
does not. Match quality is therefore expressed in \emph{durability}: high-quality
relationships are less likely to end at any given tenure.

This yields the observable counterpart of the latent object. Let
$d_{wft} \in \{0,1\}$ indicate that the relationship between worker $w$ and firm $f$
ends at tenure $t$. The separation hazard is
\begin{equation}
\lambda_{wft} = \Pr\!\big(d_{wft} = 1 \mid \text{survived to } t\big)
= \sigma\!\big(b(t) + \beta\,\MQ_{wft}\big), \qquad \beta < 0,
\label{eq:hazard}
\end{equation}
where $b(t)$ is a flexible baseline profile in tenure, common across relationships,
and $\MQ_{wft}$ is latent match quality, the counterpart of $y_{wf}=\omega_f^{\top}s_w$,
indexed by tenure because the worker's position evolves within the relationship. Two
features deserve emphasis. First, mobility frictions, search frictions, and the common
component of the outside option live in the baseline $b(t)$, which acts as the
stay-or-leave threshold, not inside $\MQ$.
Second, the sign restriction
$\beta < 0$ is a normalization, not a testable claim: only the product
$\beta\,\MQ_{wft}$ enters the likelihood, so $(\beta, \MQ_{wft})$ and
$(-\beta, -\MQ_{wft})$ fit the data identically, and the restriction fixes which of two
equivalent labelings we call match quality. What the data determine is the
\emph{magnitude} of $\beta$, an estimate of how strongly match quality governs
retention.

\subsection{Limits of the Duration Signal}\label{sec:duration-limits}

Duration is an informative but imperfect proxy for quality, in two opposing
directions. Under \emph{lock-in}, a poor match may persist because moving is costly,
so low mobility overstates quality. Under the \emph{job ladder}, an excellent match
may end precisely because it succeeded---the worker is promoted away or poached---so
high quality can coincide with short tenure. Because our estimation uses the hazard
alone, we cannot neutralize these forces inside the objective. We instead confront
them through the pattern of \emph{independent} outcomes the resulting score
predicts. Neither lock-in nor the job ladder implies that a firm whose workforce
scores highly should be more profitable or grow faster, nor that movers should
systematically choose destinations the score ranks highly. Every outcome other than
separation timing is therefore a test rather than an input
(Section~\ref{sec:strategy}); Section~\ref{sec:conclusion} returns to what the tests
do and do not establish.

\subsection{From Theory to Embeddings}\label{sec:embeddings}

The two towers give a data-driven implementation of $\omega_f^{\top} s_w$. The worker
tower maps career history, static covariates, and the worker's current position inside
the firm into a $D$-dimensional vector $u_{wt}$, a learned stand-in for the skill vector
$s_w$ as it stands at tenure $t$. The firm tower maps company characteristics into $v_f$,
a stand-in for the task-weight structure $\omega_f$. Match quality is their cosine
similarity,
\begin{equation}
\MQ_{wft} = \cos(u_{wt}, v_f),
\end{equation}
which is bounded in $[-1,1]$ and therefore scale-fixed, so that $\beta$ in
equation~\eqref{eq:hazard} is identified as a free parameter.

The cosine is not a convenience. Under the normalization adopted in
Section~\ref{sec:multidim}, $\|\omega_f\|_2 = \|s_w\|_2 = 1$, the inner product
$\omega_f^{\top} s_w$ \emph{is} the cosine, so the architecture imposes the functional
form the skill-weights model implies rather than an arbitrary similarity. What the model
is left free to learn is the content of the dimensions---which aspects of human capital
matter and how they combine---without the researcher pre-specifying them. The embedding
dimension $D$ need not equal the conceptual $K$, provided it is large enough to span the
variation the data support.

Two properties of this construction are relevant for what follows.
First, $u$ and $v$ are functions of characteristics, not identities attached to a
relationship. The score is defined for any worker and any firm whose inputs can be
assembled, whether or not that pair was ever observed. This contrasts with the estimator
a pure \citet{jovanovic_job_1979} learning model implies: a match fixed effect would provide one estimate per realized
relationship, recoverable from that relationship's own survival but silent about the
worker who was never hired. Estimating a function rather than a fixed effect is what lets
the measure price counterfactual reallocations. Second, the objective disciplines only the
cosine, so the coordinates are identified only up to a common rotation. For any
orthogonal $Q$, $\cos(Qu_{wt}, Qv_f) = \cos(u_{wt}, v_f)$, and the rotated towers fit the
data identically. What survives that rotation is the \emph{geometry}: every pairwise
similarity (among workers, among firms, and between the two) is preserved, so relative
positions, neighborhoods, and directions defined relative to the embedded points are well
defined and remain open to interpretation after the fact. What is not well defined is the
meaning of a coordinate. We therefore read the space through directions rather than through
dimensions, and attach no interpretation to individual axes.

\section{The Human Capital Transformer}\label{sec:hct}

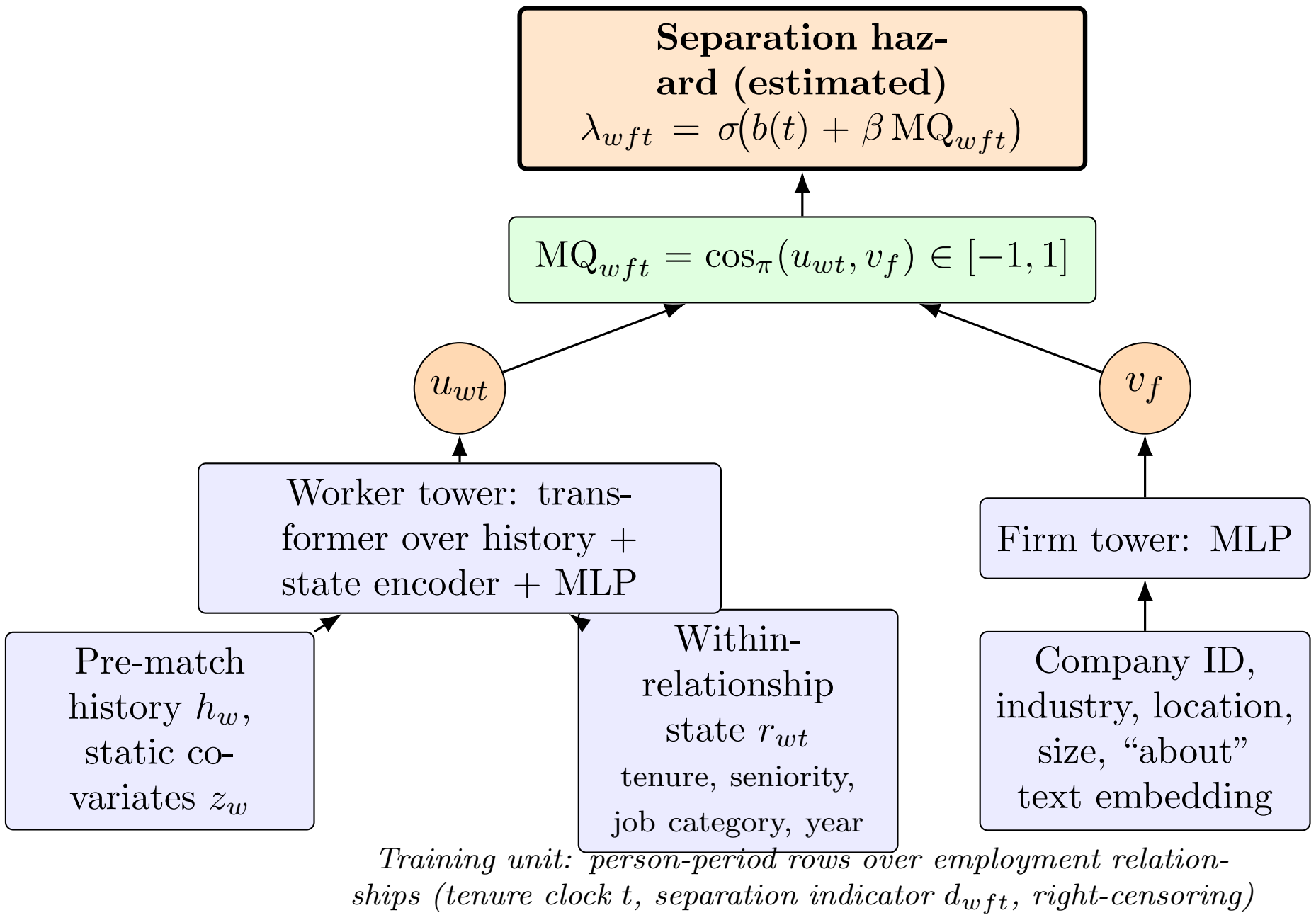
\begin{figure}[p]
\centering
\tikzset{
  box/.style={draw, rounded corners=2pt, align=center, inner sep=4pt,
              minimum height=0.75cm, fill=blue!8, font=\footnotesize},
  emb/.style={draw, circle, fill=orange!30, minimum size=0.85cm, inner sep=1pt,
              font=\small},
  score/.style={draw, rounded corners=2pt, align=center, inner sep=4pt,
                fill=green!12, minimum height=0.8cm, font=\footnotesize},
  live/.style={draw, very thick, rounded corners=2pt, align=center, inner sep=4pt,
             fill=orange!20, font=\footnotesize},
  off/.style={draw, dashed, rounded corners=2pt, align=center, inner sep=4pt,
             fill=gray!8, font=\footnotesize, text=gray!60!black},
  arr/.style={->, >=Latex, semithick},
  arroff/.style={->, >=Latex, semithick, dashed, gray}
}
\begin{tikzpicture}
  \node[box, text width=2.6cm] (wIn)    at (-2.8,0) {Pre-match history $h_w$,\\static covariates $z_w$};
  \node[box, text width=2.7cm] (wState) at (2.6,0)  {Within-relationship state $r_{wt}$\\{\scriptsize tenure, seniority, job category, year}};
  \node[box, text width=4.6cm] (wEnc)  at (0,1.8)   {Worker tower: transformer over history $+$ state encoder $+$ MLP};
  \node[emb] (wEmb)  at (0,3.2)   {$u_{wt}$};
  \node[box, text width=2.8cm] (fIn)   at (6.4,0)   {Company ID, industry, location,\\size, ``about'' text embedding};
  \node[box, text width=2.8cm] (fEnc)  at (6.4,1.8) {Firm tower: MLP};
  \node[emb] (fEmb)  at (6.4,3.2)   {$v_f$};
  \node[score, text width=5.2cm] (cos) at (3.2,4.4) {$\MQ_{wft}=\cos_\pi(u_{wt},v_f)\in[-1,1]$};
  \node[live, text width=5.0cm] (haz) at (3.2,6.0)
     {\textbf{Separation hazard (estimated)}\\$\lambda_{wft}=\sigma\!\big(b(t)+\beta\,\MQ_{wft}\big)$};
  \draw[arr] (wIn)--(wEnc); \draw[arr] (wState)--(wEnc);
  \draw[arr] (wEnc)--(wEmb);
  \draw[arr] (fIn)--(fEnc); \draw[arr] (fEnc)--(fEmb);
  \draw[arr] (wEmb)--(cos); \draw[arr] (fEmb)--(cos);
  \draw[arr] (cos)--(haz);
  \node[align=center, font=\scriptsize\itshape, text width=11cm] at (3.2,-1.4)
     {Training unit: person-period rows over employment relationships
      (tenure clock $t$, separation indicator $d_{wft}$, right-censoring)};
\end{tikzpicture}
\caption{The HCT scores a worker--firm pair as the weighted cosine of two separately
computed vectors, and that score enters a discrete-time separation hazard that is the
sole estimated objective. Both towers map into a common $D$-dimensional space.}
\label{fig:arch}

\end{figure}

This section describes the architecture and the estimated objective.
Figure~\ref{fig:arch} summarizes the design.

\subsection{Unit of Observation}\label{sec:unit}

The unit is the worker--firm \emph{employment relationship}: the continuous
association between worker $w$ and firm $f$, comprising the sequence of positions the
worker holds at that firm. Let $t$ denote tenure, normalized so that $t=0$ at the
start of the relationship. Each relationship is expanded into one row per year of
tenure, with a separation indicator $d_{wft}$ equal to one if the relationship ends
during that year.

Two-tower models are typically used in recommender systems to associate users (for
example, shoppers) with items (products to buy). In such settings, the signal the model
uses to learn is the association itself. In our context, this would mean using only transition observations, in which a
worker moves from one firm to the next.
We depart from that approach for a reason specific to employment data.
Restricting the sample to job switches
discards precisely the signal Section~\ref{sec:revelation} identifies as
informative---that workers remain in good matches longer---and it conditions the sample on
having moved. The person-period formulation keeps every worker in the sample and lets
persistence enter as \emph{duration} instead: a good match registers as a low separation
hazard at each year of tenure, rather than as a correct guess about which employer comes
next.

\subsection{Firm Tower}

The firm tower uses a shared company encoder to map each firm's fixed identity (firm ID)
and categorical features (two-digit NAICS industry code, Revelio industry classifications,
geographical information, observed size bucket) to a learned firm-specific embedding
representation $x_f$. This representation is combined with a 384-dimensional
sentence-transformer embedding of the firm's ``about'' text when available, denoted
$a_f$.\footnote{The description supplements the categorical features, since two firms
sharing an industry code can have different skill demands; firms without a description
are retained with the text component absent, as workers without a prior history are on
the worker side.} Together, they are passed through a multilayer perceptron with
$L_2$-normalization,
\begin{equation}
v_f = \phi_F(x_f, a_f) \in \mathbb{R}^{D}.
\end{equation}

\subsection{Worker Tower}

The worker tower combines three information sources. The first is the worker's
\emph{pre-match career history}
\[ H_w=\big[ (x_{f_1}, e_1), (x_{f_2}, e_2), \ldots, (x_{f_J}, e_J) \big], \]
the sequence of the worker's $J$ prior positions, each encoded by the same company
encoder that feeds the firm tower ($x_{f_j}$), together with that position's job
category and seniority ($e_j$).
This sequence is processed by a transformer encoder over at most 50
positions, and the hidden state at the most recent position, $h_w$, is taken as the
sequence summary. The history is fixed for the duration of the relationship; promotions
and role changes within the current firm enter through the within-relationship state
$r_{wt}$ described below.\footnote{
While one \emph{relationship} can contain multiple {positions}, which would appear as multiple elements
in the history, the pre-match career history $H_w$ excludes the current relationship.
This ensures that the firm identifier does not leak into the worker's vector
representation, which could lead the model to memorize worker--firm pairs.
}
The second is a vector of \emph{static covariates} $z_w$ (highest degree, predicted gender, predicted
ethnicity, detailed role category, and location) capturing time-invariant worker
characteristics.
The third is a \emph{within-relationship state}
$r_{wt}$, including current tenure, current seniority, current job category,
and calendar year.\footnote{The state is measured \emph{within} the worker's current
relationship: tenure is time since that relationship began, and seniority and job
category are those held there. Every component is nonetheless read off the worker's
own record, so the worker tower forms $u_{wt}$ without reference to firm $f$. We
index the state by $(w,t)$ to make that explicit. Strictly, $h_w$ and $r_{wt}$ are
specific to the worker's current relationship.
We suppress that index, since it is determined by $w$ and calendar time, not by $f$'s
characteristics. The restriction matters for
Section~\ref{sec:pair}: were $u$ allowed to depend on $f$, any pair-specific
variation in match quality could be encoded directly, whereas here it must arise from
the geometry of two separately computed vectors.}
The categorical features in $z_w$ and $r_{wt}$ are represented by learned embeddings.

These three components are concatenated and passed to a multilayer perceptron
with $L_2$-normalization, producing
\begin{equation}
u_{wt} = \phi_W(h_w, z_w, r_{wt}) \in \mathbb{R}^{D}, \qquad D = 128 ,
\end{equation}
the worker-side counterpart of $\phi_F$.
Including $r_{wt}$ makes the worker vector dynamic: match quality updates as
the relationship progresses through promotions and role changes, so that a worker who
is promoted and retained can register higher match quality than at the start of the
relationship, while pre-match history remains fixed.

\subsection{Match Quality}\label{sec:mq}

Match quality is the cosine similarity between the two vectors. In the implementation
the cosine is computed with learned positive per-dimension weights
$\pi \in \mathbb{R}^D_{++}$,
\begin{equation}
\MQ_{wft}
= \cos_\pi(u_{wt}, v_f)
= \frac{\sum_k \pi_k\, u_{wt,k}\, v_{f,k}}
{\sqrt{\sum_k \pi_k u_{wt,k}^2}\;\sqrt{\sum_k \pi_k v_{f,k}^2}},
\label{eq:mq}
\end{equation}
which remains bounded in $[-1,1]$ and reduces to the ordinary cosine when
$\pi = \mathbf{1}$. The weights are constrained positive through a softplus (smooth
positive) transformation, initialized at one, and estimated jointly with the
towers.\footnote{The rotational
indeterminacy noted in Section~\ref{sec:embeddings} survives the weighting. Writing
$\Pi = \mathrm{diag}(\pi)$, the weighted cosine is unchanged under $u \mapsto Au$,
$v \mapsto Av$ whenever $A^{\top}\Pi A = c\Pi$ for some $c>0$, whose solutions are
$A = \sqrt{c}\,\Pi^{-1/2} B \Pi^{1/2}$ for orthogonal $B$---a group of the same dimension
as $O(D)$. The $L_2$-normalization of the towers does not remove it, because the weighted
cosine is scale-invariant in each argument.} Economically, $\pi$ lets the model weight the
latent space unequally, so that some directions count for more than others in
determining match quality; as with the coordinates themselves
(Section~\ref{sec:embeddings}), the weights attach to directions the estimation selects
rather than to nameable aspects of skill. It is a global rather than firm-specific
weighting, so firm-specific skill demands are carried by $v_f$ rather than by $\pi$.

\subsection{Estimated Objective}\label{sec:objective}

\begin{figure}[!htbp]
\centering
\begin{minipage}[t]{.8\textwidth}
\centering
\includegraphics[width=\linewidth]{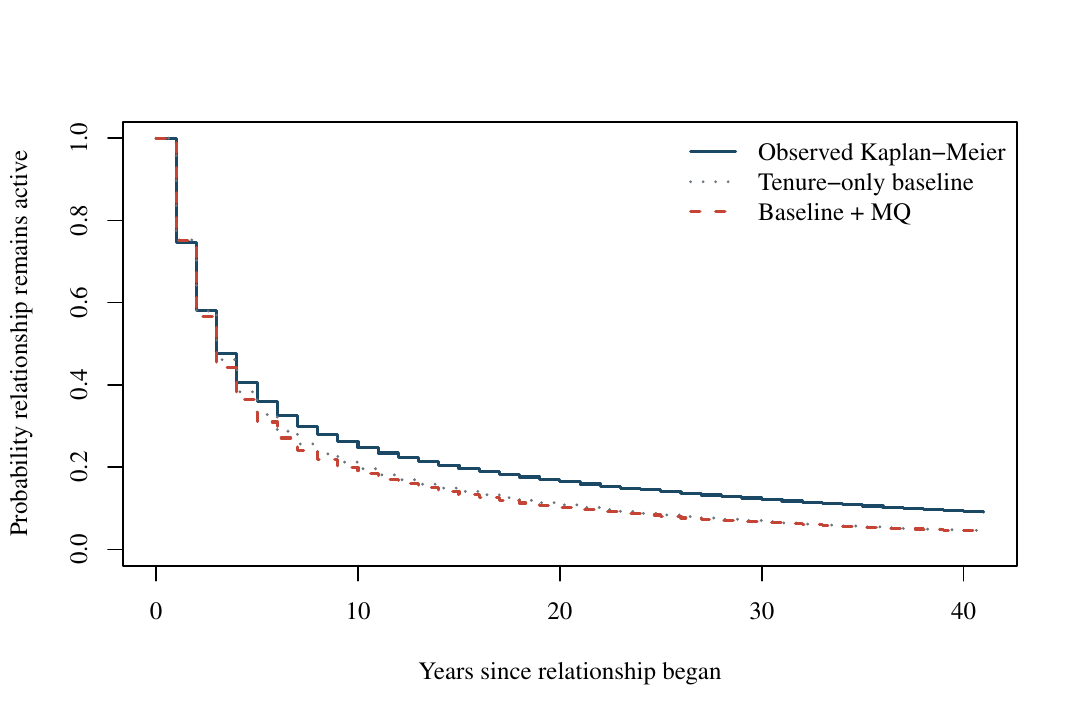}
\par\smallskip
\small (a) Survival by tenure
\end{minipage}\hfill
\begin{minipage}[t]{.8\textwidth}
\centering
\includegraphics[width=\linewidth]{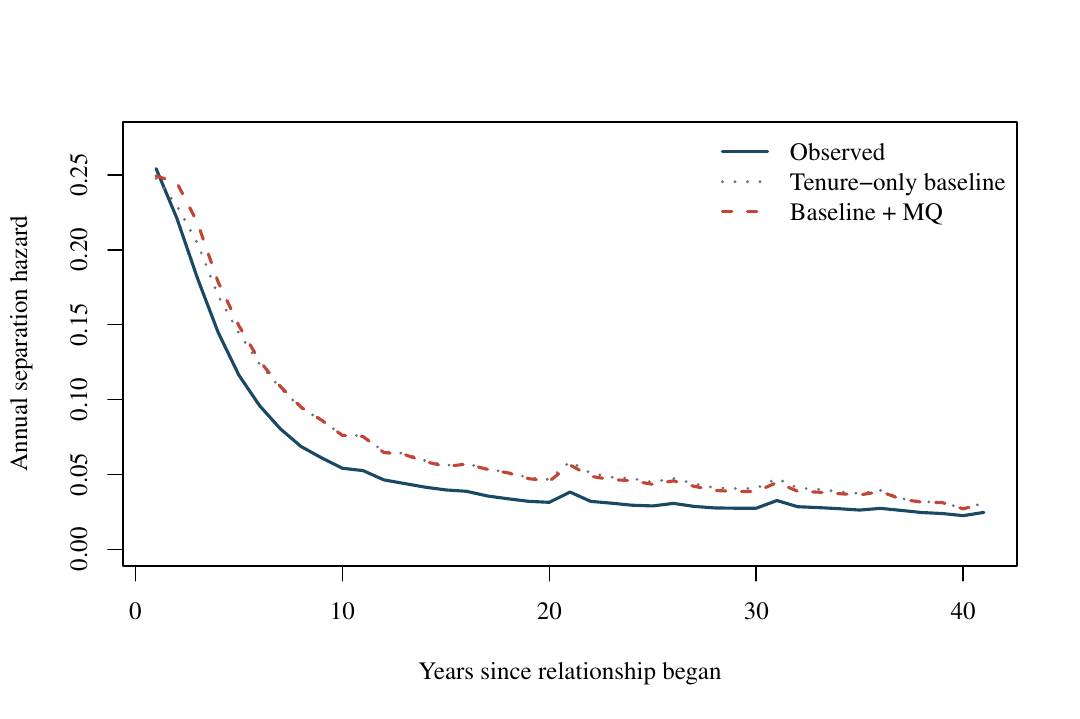}
\par\smallskip
\small (b) Annual hazard by tenure
\end{minipage}
\caption{This figure compares observed and model-implied survival and hazards by tenure.
Panel (a) compares the observed
Kaplan--Meier survivor function with the survivor functions implied by the tenure-only
baseline and by the full model (tenure baseline plus $\MQ$). Panel (b) compares the
observed annual separation hazard with the average predicted hazards from those two
specifications. Both panels use the in-sample person-periods of the full-training
checkpoint over 2008--2024.}
\label{fig:survivor}
\end{figure}

Estimation maximizes the likelihood of the discrete-time hazard in
equation~\eqref{eq:hazard}. With right-censoring, the probability of surviving through
tenure $T$ is $\prod_{t\le T}(1-\lambda_{wft})$, and the log-likelihood reduces to a binary cross-entropy
over separation indicators:
\begin{equation}
\mathcal{L}_{\mathrm{haz}}
= -\sum_{w,f,t} \Big[ d_{wft}\log \lambda_{wft}
+ (1-d_{wft})\log(1-\lambda_{wft}) \Big].
\label{eq:loss}
\end{equation}
This is the multiperiod-logit formulation of survival analysis, long used in labor
economics for job tenure and in accounting and finance for bankruptcy prediction
\citep{shumway_forecasting_2001}. The coefficient $\beta$ of equation~\eqref{eq:hazard}
is constrained negative
through a softplus transformation. The baseline
$b(t)$ is a separately estimated logistic module including an intercept
and one free parameter for each
year of tenure, up to 40. It is updated at every step alongside the towers but fit to
its own likelihood, a logit of separation on tenure alone. It is then included in the hazard
equation as a fixed offset that the tower parameters do not adjust. The baseline therefore
traces the marginal separation profile in tenure, and the towers and $\beta$ are
estimated to explain departures from that profile: $\MQ_{wft}$ carries what the
timing of separations says about a pairing beyond what its tenure alone would
predict.\footnote{The two likelihoods are summed and optimized together, the towers
and $\beta$ with a learning rate of $10^{-5}$ and the baseline with $10^{-3}$, using
AdamW with weight decay $10^{-3}$, mixed precision, and batches of 16,384
person-periods. Fitting the baseline to its own likelihood rather than jointly with the towers
is what makes $b(t)$ a benchmark the trained score can be compared against
(Figure~\ref{fig:survivor}).}

Two implementation details affect the estimand. First, rows are weighted by
employer frequency, so that a small number of very large employers do not dominate
the likelihood, making the estimator a weighted maximum-likelihood estimator.\footnote{
Each row receives a weight proportional to the inverse fourth root of
its employer's number of training rows, normalized to average one and bounded in
$[0.25, 2.0]$.
}
Second, the specification reported here includes no additive firm, worker, or relationship terms
in the hazard, so all worker-level and firm-level heterogeneity in separation
propensity must be carried by $\MQ$. Sections~\ref{sec:pair} and~\ref{sec:conclusion}
discuss what this implies for interpreting the variance decomposition.

\section{Data and Research Design}\label{sec:data}

\subsection{Data Sources}

Our primary data source is Revelio, which assembles resume-level records for
workers in the United States. For each individual we observe the sequence
of employers, position start and end dates, job categories and seniority levels,
salary estimates, inferred demographic characteristics (predicted gender and
ethnicity), education, and location. For each company we observe industry
classifications, headquarters location, an observed size bucket, and a
self-description (``about'' text) that we embed with a pre-trained sentence
transformer. Salaries in Revelio are modeled rather than observed, which is why we
use them only as validation targets and never as model inputs.

We link firms to Compustat and CRSP for the subsample of publicly traded companies, which
provides total assets, revenue, employment, R\&D expense, market value, and
other controls used in Section~\ref{sec:firm}. We additionally link the
patent data of \citet{kogan_technological_2017} for patent counts, citations, and
patent market values. Finally, we use the purchase-price allocations reported in XBRL for the
goodwill recorded on acquisitions, over the period covered by the model.
None of these sources enters model estimation.
Appendix Table~\ref{tab:vardef} defines every variable used in the analysis.

\subsection{Sample Construction}

We restrict the sample to workers located in the United States. A Revelio profile
lists the positions a worker has held, each with an employer, a job title, and start
and end dates. We group the consecutive positions a worker holds at
the same employer into a single employment relationship, so that a promotion or a
change of role within a firm continues the relationship. The
change is still recorded: it updates the worker's within-relationship state. We then
follow each relationship year by year to create \emph{person-period} observations, one
for each year of the worker's tenure, recording whether the worker left the firm that year
(Section~\ref{sec:unit}). Tenure is capped at 40 years, and the observation window
runs from 2008 to 2024.\footnote{Relationships that began before 2008 are retained with their
true start dates, so that tenure and pre-match history are measured correctly, but
contribute observations only from 2008 onward.
} Table~\ref{tab:sample} describes the resulting
sample. It comprises
100.6 million workers, 8.4 million firms, 277.4 million worker--firm relationships,
340.7 million employment positions, and 1.18 billion person-periods. The median
completed relationship lasts 1.83 years and the mean 3.26 years; a next employer is
observed for 87\% of separations; 86\% of firms have usable ``about'' text. Among
workers with a completed relationship, the average is 3.03 relationships. The annual separation rate in the risk set is 14\%.

Workers with no career history before the start of a relationship are retained: the
history encoder accepts an empty sequence, so a worker in a first job enters through
static covariates and the within-relationship state alone, and the sample is not
tilted toward workers observed beyond their first job. The firm-level analyses use
Compustat-linked firms over 2008--2024: of
163,331 Compustat firm-years (20,654 firms), 94,459 (10,064 firms) carry a match-quality
score, 71,873 (7,705 firms) have at least 20 scored relationships, and 48,432 (5,623
firms) have the controls and outcomes the regressions require (Appendix
Table~\ref{tab:firmsample}; Appendix Table~\ref{tab:firmdescriptivestats} reports
descriptive statistics for the firm-level samples).

\subsection{Splits, Checkpoints, and Analysis Samples}

Splits are assigned at the \emph{worker} level using a stable seeded hash of the
worker identifier, with 60\% of workers in training, 20\% in validation, and 20\% in
test. All person-periods for a given worker fall in the same split. ``Out of
sample'' in what follows therefore means \emph{held-out workers}: the model is
evaluated on people it has never seen, over the same period covered by training. Three
checkpoints appear in the paper. The \emph{full-training} checkpoint is estimated on all
splits; Table~\ref{tab:hazest} reports its parameters. The \emph{training} checkpoint is
estimated on the training and validation splits and is the one evaluated on held-out
workers (Table~\ref{tab:hazperf}, Table~\ref{tab:pair}, and the
destination test of Section~\ref{sec:dest}). Because in-sample and held-out performance are close, the
full-training scores serve the in-sample worker-level tables (Tables~\ref{tab:hazest}
and~\ref{tab:wage}) and the
firm-level tests (Tables~\ref{tab:firm}--\ref{tab:ma}). A third, \emph{time-split}
checkpoint is estimated on data through 2016 only and serves the tests of Appendix
Tables~\ref{tab:firmhor2016}, \ref{tab:innovhor2016}, and~\ref{tab:ma2016}, which
relate match quality measured in 2016, the last year that checkpoint saw, to outcomes
from 2017 onward; those are the only tests in which the model is evaluated on years it
was not trained on.

The two worker-split checkpoints are estimated on the full Revelio universe, subject to
the split above. The in-sample worker-level tables (Tables~\ref{tab:hazest} and~\ref{tab:wage})
are computed on a five-percent
random subsample of workers drawn from the scored file---58.8 million person-periods,
4.5 million workers, and 1.9 million firms---which is the largest the post-hoc
fixed-effects regressions accommodate. Table~\ref{tab:hazperf} uses the held-out workers
within that subsample; Table~\ref{tab:pair} uses the training checkpoint on the same
subsample, with the training and validation workers as its in-sample columns and the
held-out workers as its out-of-sample columns; Table~\ref{tab:dest} uses held-out
separations; and the firm-level tables use every Compustat-linked firm-year.

\subsection{Empirical Predictions and Validation Strategy}\label{sec:strategy}

The framework of Section~\ref{sec:theory} yields five empirical predictions, which
organize Section~\ref{sec:results}. Sections~\ref{sec:hazest}--\ref{sec:pair} test the
first two (the estimated coefficient and its stability, the out-of-sample test, and the
variance decomposition), Sections~\ref{sec:wages} and~\ref{sec:dest} the third and
fourth, and Sections~\ref{sec:firm}--\ref{sec:ma} the fifth. The first two concern the measure itself:
whether $\MQ$ predicts separations out of sample, and whether that relation is a
property of the worker--firm pair. The other three concern outcomes that never enter
estimation, and so validate the measure outside its objective. Throughout, we interpret the estimates as associations rather than
causal effects.

\paragraph{Separation prediction.} The first prediction is that $\beta < 0$ and that
$\MQ$ carries information about separations beyond duration dependence and
observables. Because $\beta$ is a parameter of the objective that maximizes precisely
this fit, its magnitude alone is not evidence; the evidence is out of sample. We
report two standard measures of predictive performance on held-out workers---the AUC
and cross-entropy---together with in-sample calibration by tenure,\footnote{The AUC (area under the receiver
operating characteristic curve) is the probability that the model assigns a higher
separation probability to a randomly chosen relationship that ends than to one that
continues; 0.5 is a coin flip and 1 is perfect ranking. Cross-entropy measures how
close the predicted probabilities are to what actually happened: it averages $-\log$
of the probability the model assigned to the observed outcome, so it is lower when the
model is both right and confident, and it is the same quantity the model minimizes in
estimation (equation~\eqref{eq:loss}). Calibration asks whether predicted separation
rates match realized ones on average.} and compare the model against a ladder of
conventional baselines that adds, in turn, the
tenure profile, calendar-year effects, and observable worker and firm characteristics.
Finally, we enter $\MQ$ as a single covariate in that conventional hazard
(Table~\ref{tab:hazperf}); Figure~\ref{fig:survivor} reports the calibration.

\paragraph{Stability and pair specificity.} The second prediction is the framework's
sharpest restriction and the one that separates it from existing measures: the
association between $\MQ$ and separations reflects the pairing itself, not aggregate
conditions, observable composition, or additive worker and firm quality. Were skill
one-dimensional (Section~\ref{sec:multidim}), $\MQ$ would be additively separable into a
worker effect and a firm effect. Worker-level, firm-level, and occupation-level measures
impose that separability by construction; the skill-weights model requires it to be
false. By design, $\MQ$ can also reflect the business cycle and the composition of
who works where: calendar year is an input to the worker tower while the baseline
$b(t)$ depends on tenure alone, and the towers see worker and firm observables. We do
not claim that $\MQ$ is free of these components, only that they do not drive its
association with separations. We test this in two steps. First, holding the trained
score fixed, we add calendar-year fixed effects and then the observables to a
separation logit (Table~\ref{tab:hazest}); if they accounted for the association, the
coefficient on $\MQ$ would shrink substantially. Second, a variance decomposition
measures how much of the variation in $\MQ$ is left after removing additive worker and
firm effects, in sample and on held-out workers, both scored by the training checkpoint
(Table~\ref{tab:pair}). A sizable residual is necessary but
not sufficient, since part of it could be noise in the score; the outcomes below, none
of which entered estimation, are designed to show whether it carries economic content.

\paragraph{Construct validity: priced and acted upon.} If $\MQ$ measures the value
of a pairing, then agents other than the econometrician should respond to it. The
third prediction is that higher match quality is associated with higher wages,
conditional on observables and, in the demanding specification, within worker and
within firm, since in equilibrium workers appropriate part of the surplus
\citep{jovanovic_job_1979, topel_specific_1991}; the association should be strongest
early in a relationship, while the surplus is being revealed and priced
(Table~\ref{tab:wage}). The fourth prediction is that the score should order
\emph{unrealized} pairs, not merely rate realized ones. Because $\MQ$ is a function of
worker and firm inputs rather than an identity attached to a relationship
(Section~\ref{sec:embeddings}), it is defined for pairs that never formed, so when a
relationship ends the realized destination should rank highly among a thousand
candidate employers, a ranking task the model was never trained on
(Table~\ref{tab:dest}). This is the prediction a match fixed effect cannot state, and
it is the test that distinguishes a measure of compatibility from a measure of
inertia: a score that merely tracked who stays would predict neither what a worker is
paid nor which firm the worker joins next.

\paragraph{Aggregation.} The fifth prediction is that a measure built from
individual relationships aggregates: firms whose workforces are better matched should
perform better, because the productivity of a well-matched workforce translates into
higher output and lower coordination costs, and because innovation is produced by
people, they should also innovate more, with the lag that patenting entails
(Tables~\ref{tab:firm}--\ref{tab:innov}). The fifth prediction extends to acquisitions:
because $\MQ$ is defined for pairs that do not yet exist, the fit of a target's workers
with the acquirer, relative to their fit with the target, should be reflected in the
share of the price the acquirer records as goodwill (Table~\ref{tab:ma}).

\section{Empirical Findings}\label{sec:results}

\subsection{Hazard Estimates and Their Stability}\label{sec:hazest}

Table~\ref{tab:hazest} reports the coefficient on match quality in the separation
hazard. The first column is the HCT as estimated (full-training
checkpoint): $\hat{\beta} = -6.49$.
Because $\MQ$ is bounded in $[-1,1]$ and therefore scale-fixed, the coefficient has a direct reading. A 0.1 increase in match
quality multiplies the odds that a relationship ends in a given year by
$e^{-0.649} = 0.52$, a reduction of 48\%; a one-standard-deviation increase (about 0.13
across person-periods) multiplies them by 0.42. In levels, at the sample's annual
separation rate of 14\%, a relationship 0.1 higher in match quality separates with a
probability of about 7.8\%, and one a full standard deviation higher with a
probability of about 6.4\%.

Columns~(2) and~(3) ask whether the coefficient mainly reflects something other than
pair-level match quality. Because calendar year enters the worker state $r_{wt}$ while
$b(t)$ depends on tenure alone, part of the cyclical variation in separations is carried
by $\MQ$ (Section~\ref{sec:strategy}). Both columns are logits estimated after training on
the scored person-period file, with $\MQ$ held fixed at its trained value; nothing is
re-estimated. Column~(2) adds calendar-year fixed effects to the tenure baseline. The
estimate is $-6.49$ against $-6.46$ in the tenure-only logit of column~(1) ($p < 0.001$
in both). Column~(3) adds the worker and firm observables the towers see---highest
degree, predicted gender and ethnicity, detailed role category, and location on the
worker side; industry classifications, headquarters metropolitan area and country, and
size bucket on the firm side---together with current seniority and job category
(Appendix Table~\ref{tab:vardef}), and gives $-6.51$ ($p < 0.001$). Across the three logit specifications the
coefficient moves by less than 1\%, and all three lie within half a percent of the trained value. $\MQ$ does capture cyclical and compositional
variation, but that is not what drives its relation to separations: the variation that
remains once calendar time and observable composition are absorbed carries essentially
the whole coefficient.

One thing should be kept in mind when reading the magnitude. $\beta$ is a parameter of
an objective that maximizes precisely this fit. Its size records how much of the timing
of separations a single pair-level scalar can absorb when the towers are free to
construct that scalar from 1.18 billion person-periods; it is not the response of
turnover to an exogenous change in match quality, and we do not read it as one.
Instead, the coefficient sets up the sequence of questions the rest of this section asks:
whether the scalar carries information the tenure profile does not
(Table~\ref{tab:hazest}), including on workers the model has never seen
(Table~\ref{tab:hazperf}); whether it is a property of the pair rather than of the
worker or the firm (Table~\ref{tab:pair}); and whether it predicts outcomes it was
never fit to (Tables~\ref{tab:wage}--\ref{tab:ma}).

Figure~\ref{fig:survivor} presents the fit in the form the duration literature uses:
Panel~(a) compares the Kaplan--Meier survivor function of observed relationships with
the survivor functions implied by the tenure baseline alone and by the full model,
$\prod_{s \le t}(1-\hat{\lambda}_s)$ on the same risk set, where $\hat{\lambda}_s$ is the
average predicted hazard at tenure $s$; Panel~(b) does the same for the annual hazard.
The negative duration dependence is the pattern \citet{jovanovic_job_1979} predicts and
that \citet{farber_mobility_1999} documents for U.S. jobs, and it is recovered here
without being imposed: $b(t)$ is estimated as a free profile. Adding $\MQ$ moves both
the implied survivor function and the annual hazard toward the observed series at every
tenure, although both specifications predict somewhat more separation than observed
beyond the first few years. That is the in-sample calibration evidence.

For the match-quality coefficient itself, the nearest points of comparison are
\citet{fredriksson_mismatch_2018} and \citet{hoffman_discretion_2018}. The former
measure mismatch as the distance between a worker's cognitive and non-cognitive skill
profile and that of their coworkers and find that it raises separations, most strongly
for inexperienced workers about whom less is known at hiring; the latter find that
hires recommended by a job test stay longer than those the test rated poorly. Both
measure a fixed attribute of the worker or the hire; ours is a score fit to
separations themselves, which is why we treat the out-of-sample and pair-specific
tests, not the coefficient, as the evidence. How well the model tells apart
relationships that end from those that continue is the subject of the next table.

\subsection{Does Match Quality Predict Separations?}\label{sec:oos}

Table~\ref{tab:hazperf} reports the central out-of-sample validation of $\MQ$. The question is
whether $\MQ$---the pair-level score---carries information about separation beyond
what a conventional hazard already has, on workers the model has never seen. The table
is a ladder. It begins with a constant separation rate and adds, in turn, the tenure
profile, calendar-year effects, and the worker and firm observables of
Table~\ref{tab:hazest}; the last logit adds the trained $\MQ$ as one more covariate, so
that the ladder asks whether $\MQ$ adds to worker characteristics, firm
characteristics, and aggregate conditions rather than merely restating them. Every
model is fit on training and validation workers and evaluated on the held-out workers
(the 20\% test split) within the five-percent analysis subsample (11.7 million
person-periods), over the same calendar period as training.\footnote{Logit rows can use slightly fewer
held-out person-periods when a fixed-effect category in the held-out data is absent
from the training data.}

Duration dependence alone is a theoretically motivated benchmark: the tenure profile
raises the AUC from 0.500 to 0.695, for the HCT's tenure baseline $b(t)$ and a logit
with tenure fixed effects alike, and lowers cross-entropy from 0.403 to 0.373 (0.376
for the HCT's own baseline). Calendar-year effects add a further 0.012 and the
observables a further 0.027, so the richest conventional specification---the Panel~B
logit with tenure, calendar year, and controls---reaches an AUC of 0.734. The HCT with
$\MQ$ in Panel~A reaches 0.764. The gain of 0.030 over that logit---and of 0.069 over
the HCT's own tenure baseline, with cross-entropy falling from 0.376 to 0.348---is the
main result: a single scalar summary of the pair,
formed from career history, firm characteristics, and the worker's current position,
improves the prediction of which relationships end, on held-out workers, by more than
calendar time and observable composition together.

The last row of Panel~B separates the content of $\MQ$ from the functional form that
produced it. Entering the trained score as a covariate in the same logit---no
transformer, no towers at prediction time, one coefficient---raises the AUC from 0.734
to 0.778, above the HCT's 0.764, and cross-entropy falls to 0.339,
below the HCT's 0.348 in Panel~A. All of the HCT's advantage over the conventional
specification therefore survives being reduced to a linear term in a logit, which is
the form a reader could estimate with the score file alone; the logit does somewhat
better because it also absorbs the calendar-year effects and observables that the
HCT's tenure-only baseline omits. The gain lies in the score, not in the neural
network that produced it.\footnote{The full-training checkpoint's in-sample McFadden
pseudo-$R^2$ is 0.152 (Table~\ref{tab:hazest}), against 0.138 for the training
checkpoint on held-out workers (Table~\ref{tab:hazperf}), so the gap between fit and
held-out performance is small; the model is not memorizing individual careers, which
is what justifies using the full-training checkpoint for the in-sample worker-level
and firm-level analyses (Section~\ref{sec:data}).} The table does
not yet separate the part of that information that belongs to the pair from the part
that belongs to the worker or the firm alone.

\subsection{Is Match Quality Pair-Specific?}\label{sec:pair}

Table~\ref{tab:pair}, the core of the paper, tests the restriction
Section~\ref{sec:multidim} derived from the skill-weights model. It decomposes the variance of $\MQ$ across worker and firm
dimensions by regressing the pair-level score on worker fixed effects, on firm fixed
effects, and on both entered additively. The additive specification deliberately
contains no worker--firm term, which would absorb the object being measured.

In sample, across 9.5 million scored pairs formed by the workers the model was trained
on, the standard deviation of $\MQ$ is 0.137.
Worker fixed effects alone explain 49.8\% of its variance and firm fixed effects alone
44.1\%; entered together they explain 73.0\%. The remaining 27.0\% is pair-specific: it
cannot be attributed to some workers being generally well matched or to some firms
generally matching their workers well. The residual has a standard deviation of 0.070,
about half the standard deviation of $\MQ$ itself. The out-of-sample columns
repeat the decomposition on 2.4 million pairs formed by held-out workers and scored
with the same training checkpoint, which never saw them. There the standard deviation of
$\MQ$ is also 0.137; worker fixed effects alone explain 49.7\% of its variance, firm fixed
effects alone 46.6\%, and the two together 74.9\%, so the pair-specific share is
25.1\%, with a residual standard deviation of 0.067. The pair component is therefore
not an artifact of pairs the model has fitted. Worker effects are the largest single
component, but about a quarter of the score belongs to neither the worker nor the firm
alone.

Economically, this is the sense in which the measure is context-dependent. The same
worker is a materially better fit for some employers than for others, by an amount
about half the dispersion of the score across all pairs, and neither a ranking of
workers nor a ranking of firms recovers it.

One caveat applies. Because the reported
hazard contains no additive worker or firm terms, $\MQ$ is the only channel through
which worker-level and firm-level separation propensity can enter the model, so the
score is forced to carry main effects that a richer specification would absorb
elsewhere; the pair share is a lower bound on that account.\footnote{Fixed-effects
decompositions of this kind are also subject to limited mobility bias. In wage
regressions with worker and firm effects, the firm effects are identified only from
workers who move between firms; when movers are few, the effects are estimated
imprecisely, and the sampling error is counted as variance explained by the additive
terms \citep{andrews_high_2008, kline_leaveout_2020, bonhomme_how_2023}. A related
problem applies to Table~\ref{tab:pair}: a worker observed in a single pair is fitted
exactly by their own effect, so the additive $R^2$ is overstated and the pair share
understated, which is a second reason the share is conservative. The measure itself is
less exposed than a wage decomposition. Match quality carries no worker identity, and
the firm vector, which includes a learned firm identifier, is estimated from the
survival of all of the firm's relationships rather than from the wages of the workers
who move between firms, so the score does not require a connected set of firms and is
defined for pairs that never formed; the
fixed-effects regression here is a descriptive check on the score, not part of its
construction.} A variance share is
descriptive: it says the residual exists, not that it matters. Tables~\ref{tab:wage} and~\ref{tab:dest} ask whether it is also priced and acted upon.

\subsection{Match Quality and Wages}\label{sec:wages}

Table~\ref{tab:wage} reports associations between $\MQ$ and log wages. Coefficients are
expressed as the percentage wage change associated with a 0.1 increase in match quality,
with standard errors two-way clustered by worker and firm. Wages never entered
estimation, so whatever the table shows is a prediction of the model rather than a moment
it was fit to; and because Revelio salaries are modeled rather than observed, we read
the table for pattern and sign rather than as a precise quantitative exercise. The
estimates use the full-training checkpoint in sample.

Panel~A pools all relationships. The unconditional association is 7.9\%. Adding
calendar-year and tenure-year fixed effects, current seniority, and the worker and firm
observables of Table~\ref{tab:hazest} lowers it to 3.4\%. Part of the raw association
therefore runs through the observable correlates of pay---the well-matched are somewhat
more often the workers whose degrees, roles, and employers command high wages---but
more than two fifths of it survives, so match quality is not a proxy for those correlates.
Adding firm fixed effects, so that the comparison is between workers at the same
employer, gives 2.9\%; adding worker fixed effects instead, so that the comparison is
within a worker across employers and years, gives 2.1\%. The
specification with both is the match-specific wage test: with additive worker and firm
heterogeneity absorbed on both sides, what remains of the wage is its worker--firm
component and what remains of $\MQ$ is its pair component. That coefficient is 1.3\%
per 0.1 of $\MQ$, or about 1.8\% per standard deviation of $\MQ$
across pairs (0.137).

Panels~B and~C split the sample at five years of tenure and locate where the
association lives. The pattern is the same in every column: the association is larger
for relationships at tenure 0--4 (their first five years) than for those at tenure 5 or
more. The wage premium per 0.1 of match quality is 11.1\% early against 1.3\% late
unconditionally, 3.7\% against 2.8\% with observables, 3.6\% against
1.9\% within firm, 2.6\% against 1.2\% within worker, and 2.3\% against 0.1\% under
worker and firm fixed effects together. The gap between early and late tenure, not the level in either group,
is the robust feature of the table: it holds whether the comparison is across workers,
within a firm, within a worker, or within a pair, and in the last specification the
long-tenure coefficient is indistinguishable from zero. The pair
surplus is priced early, while the
value of a pairing is being revealed and acted upon, and the match-specific premium
fades as relationships age.\footnote{\label{fn:fading}Besides the lock-in discussed
below, two forces would produce the fading. First, learning (Section~\ref{sec:revelation}): by the fifth year
the quality of the match is largely known to both sides and has been priced, so later
wage growth tracks firm-specific human capital and seniority-based pay rather than fit
\citep{jovanovic_job_1979, topel_specific_1991}. Second, selection: relationships that
survive five years are disproportionately good matches, which compresses the variation
in $\MQ$ left to be priced.}

Two remarks on interpretation. First, the two-way fixed-effects column identifies from
within-relationship movement in $\MQ$, which arises from the worker state $r_{wt}$:
promotions and role changes move the worker vector while pre-match history is fixed.
That a worker who is promoted and retained is scored as better matched, and is paid
more, is what the framework predicts; but modeled salaries also load on title and
seniority, so some of the within-pair association may be mechanical, which is one
reason we lean on separations, destinations, and firm performance rather than on wages
alone. Second, the gap between early and late tenure suggests that, at long tenure,
match quality also captures lock-in: some relationships last because moving is
costly---long-tenured workers have fewer outside options and sit inside pay structures
that adjust slowly---so a high score need not command higher pay. Because learning and
selection predict the same gap (footnote~\ref{fn:fading}), the wage evidence points to a
lock-in component without establishing its size; the destination and firm-level tests
below bear on it more directly.

\subsection{Where Do Movers Go?}\label{sec:dest}

If $\MQ$ measures compatibility rather than inertia, then when a relationship ends the
worker should move to firms the measure ranks highly. Table~\ref{tab:dest} tests this
on held-out separations with an observed next employer, comparing the realized
destination against 999 uniformly sampled, time-valid alternative firms, that is, firms
observed as active employers in the year of the move. The worker vector is computed once
from the worker's own record and held fixed while the firm vector varies, so every
candidate is scored as the pair it would form.

Against the candidate pool (Panel~A), the model ranks the realized destination
first among the 1,000 candidates for 0.68\% of moves, against 0.10\% under random
ranking, and in its top ten for 3.4\%, against 1.0\%: ratios to the random rate of 6.8
and 3.4. The
mean reciprocal rank is 0.019, against 0.0075. The lift does not come from recognizing
the worker's industry. Movers who switch broad industry are ranked slightly better than
movers overall (Recall@10, the share of moves whose destination ranks in the top ten, of
3.6\%), and better than movers who stay in their industry
(2.4\%). Panel~B restricts the candidates to the destination's own industry. Because it
takes the industry the worker ends up in as known, it is not a forecast; it asks only
whether, within that industry, the score separates the firm the worker joins from its
competitors. It does: the realized destination is in the top ten for 6.5\% of moves,
more than six times the random rate.

These magnitudes are modest in absolute terms, and they should be. People also move for
pay, geography, family, and displacement, and most of the candidate firms were not
hiring when the worker moved. The result establishes direction rather than determinism:
match quality shifts the distribution of destinations, and it does so having never been
shown a destination during estimation. This is the paper's test of choice behavior, and
it is consistent with the matching view of \citet{jovanovic_job_1979}: movers do not
search randomly; the
firms they join tend to be ones the score ranks as better fits.

\subsection{Firm Performance}\label{sec:firm}

Table~\ref{tab:firm} aggregates $\MQ$ to the firm-year level for Compustat firms.
\emph{Firm match quality} (Firm MQ in the tables) is the mean of $\MQ_{wft}$ over the
firm's scored employment relationships in the year, for firms with at least 20; in the
firm-level analyses, $t$ indexes fiscal years rather than tenure. The table relates firm
match quality to growth and profitability in the following year; Appendix
Tables~\ref{tab:firmhor}--\ref{tab:firmhorfe} extend it to horizons of two to five
years. Firm match quality enters in its original units, so coefficients are per unit of
$\MQ$ and a 0.1 increase corresponds to a tenth of each coefficient. The outcomes are
next-year log revenue growth, log employment growth,
growth in revenue per employee, and the change in return on assets; the appendix
outcomes are the same, annualized from $t$ through $t+2$ to $t+5$ (for return on
assets, average ROA over the window minus $ROA_t$). All columns control for the
outcome's level and lagged change, the standard deviation of $\MQ_{wft}$ across the firm's
scored relationships, the
log number of scored relationships, log current employees, R\&D intensity with an
indicator for missing R\&D, log assets, log market value, book-to-market, leverage,
tangibility, an M\&A indicator, and return volatility; continuous variables are
winsorized at the first and ninety-ninth percentiles, and standard errors are
clustered by firm. Column~(1) includes industry and year fixed effects; column~(2), our
preferred cross-firm specification, uses industry-by-year fixed effects with
three-digit NAICS industries; and column~(3) adds firm fixed effects. The next-year
sample is 48,336 firm-years over 2008--2024 (47,775 with firm fixed effects), after
dropping singleton fixed-effect cells; Appendix Table~\ref{tab:firmdescriptivestats}
reports descriptive statistics. Every specification compares firms within
year, because firm match quality trends over calendar time (calendar year is an input
to the worker tower).

Across firms within an industry and year (column~(2)), a 0.1 increase in firm match
quality is associated with next-year revenue growth 5.4 percentage points higher,
employment growth 4.2 points higher, and a rise in return on assets 1.4 points larger;
since firm match quality has a standard deviation of 0.07 across firm-years (Appendix
Table~\ref{tab:firmdescriptivestats}), a one-standard-deviation increase corresponds
to revenue growth about 3.8 points higher.
Revenue per employee grows faster as well, but by much less (0.8 points), so
across firms better-matched firms grow mainly by scale, with revenue and headcount
expanding together. Within firm, the productivity association is larger (2.7 points).
The associations persist over longer windows (Appendix
Table~\ref{tab:firmhor}), and the profitability association grows modestly with the
horizon, from 1.4 points next year to 1.9 at three years and 2.0 at five. Within firm
(column~(3) and Appendix Table~\ref{tab:firmhorfe}), the growth associations remain
large: a firm whose workforce becomes better matched grows faster than it did when it
was worse matched, by 7.0 points in revenue growth and 4.2 in employment growth the
next year, and by 4.6 and 3.1 points a year over five years. The within-firm
association with profitability appears only with a lag: it is small and not
statistically significant the next year (0.5 points) but 1.2 points at two years, 1.9
at three, and 2.6 at five. Within a firm, an
improvement in match quality takes several years to show up in profitability, as one
would expect if a better-matched workforce needs time to translate into margins. The
lag parallels \citet{edmans_does_2011}, who finds that the value of employee-related
intangibles is reflected in earnings and prices only gradually.
Appendix Table~\ref{tab:firmhor2016} uses the time-split checkpoint, trained on data
through 2016 only, fixes match quality and the controls in that single year, and
measures the outcomes at horizons of one to five years, 2017 through 2021, so that no
outcome overlaps the years the model was trained on. The growth associations persist at
every horizon in that cross-section; the profitability and productivity coefficients
are positive but not statistically significant in that smaller sample.

Three features of the design matter for what the associations mean. First, market
value and book-to-market are among the controls, so the result is not that well-matched
firms are highly valued firms: match quality is associated with growth and profitability
holding valuation fixed. Second, every outcome is measured after the year in which
match quality is measured, so the association is not driven by good years attracting
or retaining well-matched workers within the same year. Third, and most important, no
financial variable entered estimation. The score was fit to the timing of worker
separations in a sample overwhelmingly composed of private firms, and it nonetheless
orders public firms by growth and profitability within industry and year.

\subsection{Innovation}\label{sec:innovation}

Table~\ref{tab:innov} relates the firm match quality of Section~\ref{sec:firm} to the
patents issued to the firm from $t+1$ through $t+5$, to the citations those patents
receive, and to their real market value following \citet{kogan_technological_2017}.
The five-year window reflects that innovation takes time: work that begins in a firm
this year surfaces as a patent application only after the invention is complete, as a
granted patent after examination, and as citations and market value only as later
inventors and investors respond. The lag from input to measurable output is
conventionally put at two to three years \citep{hall_patents_1986, hall_nber_2001}, so
studies of firm-level innovation read patent outcomes over windows that begin several
years after the event of interest \citep{bernstein_going_2015,
kogan_technological_2017}; five years is the window \citet{bernstein_going_2015} uses
to study how innovation changes after a firm goes public. We use issued (granted)
rather than filed patents, since a grant is the patent office's judgment that an
application is a genuine invention. Patent counts are estimated by Poisson
pseudo-maximum likelihood, which keeps firm-years without patents; citations and value
per patent are estimated by OLS on firm-years with at least one patent. Appendix
Tables~\ref{tab:innovhor}--\ref{tab:innovhorfe} report the same outcomes cumulated
through each horizon from $t+1$ to $t+5$. Controls and fixed effects are as in
Table~\ref{tab:firm}, match quality is in original units, and standard errors are
clustered by firm; because the dispersion of match quality within a workforce turns
out to matter for innovation, the tables also report the coefficient on its standard
deviation across the firm's scored relationships, which Table~\ref{tab:firm} treats as
a control. Patent data are truncated near the end of the sample as later applications
are still pending \citep{lerner_seru_2022}; we take 2024 as the last usable patent
year, so for the five-year window match quality and controls are measured through
2019.

Across firms within an industry and year (column~(2)), better-matched workforces are
issued more patents: a 0.1 increase in firm match quality is associated with about 16\%
more patents issued over the five years from $t+1$ (a coefficient of 1.45), and the
estimate is the same with industry and year effects entered
separately (column~(1)). The horizon matters, as the lag argument implies it should.
The coefficient rises steadily from 0.39 for patents issued in $t+1$, indistinguishable
from zero, to 0.68 at two years and 1.45 at five (Appendix Table~\ref{tab:innovhor}).
Within firm (column~(3)), the count association follows the same path, from essentially
zero at $t+1$ to 1.41 at five years, where it is significant at the 10\% level
(Appendix Table~\ref{tab:innovhorfe}); the within-firm estimates are less precise, so
the quantity result is sharper across firms than within them.

The patents are also better. Patents
issued to better-matched workforces are more cited across firms at every horizon, and
within firm for the earliest patents.
The association is strongest for the earliest patents: at $t+1$, a 0.1 increase in
match quality is associated with about 9\% more citations per patent across firms
(0.88) and 10\% within firm (0.93). It then
weakens steadily as the window lengthens, to 0.43 across firms over five years,
significant at the 10\% level, and to 0.08 within firm, indistinguishable from zero
(Appendix Tables~\ref{tab:innovhor}--\ref{tab:innovhorfe}).
The weakening is what citation truncation produces: patents issued late in a five-year
window have had less time to be cited by the end of the data, so a longer window
dilutes the measure with recent patents whose citations have not yet accrued. The same
patents are also worth more: a 0.1 increase in match quality is
associated with about 18\% higher real value per issued patent over five years across
firms (1.69) and 19\% within firm (1.71), an association that grows with the window in
both comparisons. Appendix Table~\ref{tab:innovhor2016} repeats the cross-firm
specification with the time-split checkpoint, match quality and controls fixed in 2016,
and the outcomes cumulated from 2017 through each year to 2021: the count and value associations appear from the
second and third years on, and the citation association at every horizon.

The dispersion of match quality within a workforce enters alongside its mean, and its
coefficient is positive and significant for the quality outcomes: across firms, for
citations per patent at every horizon (2.26 over five years) and for real value per
patent from three years on (2.69 over five years), and within firm for real value per
patent at four and five years (1.75 over five years). For patent counts it is never
significant, and within firm it is mostly negative. We read this cautiously: holding the average fit
of the workforce fixed, a workforce whose members fit the firm to differing degrees
produces, if anything, better rather than more patents, which is what one would expect if invention draws on a
mix of skill profiles rather than on uniform fit. Diversity within organizations has long
been thought to trade coordination against creativity, and
\citet{corritore_duality_2020} show that the form of heterogeneity matters: cultural
heterogeneity across a firm's employees is associated with lower profitability, while
heterogeneity within individuals is associated with more patenting. Dispersion in fit
across a workforce is closer to the former, so its positive association with innovation
here is suggestive rather than settled.\footnote{The design features discussed for Table~\ref{tab:firm} apply here as well, including
the control for R\&D intensity, so the result is not that firms that spend on research
happen to have well-matched workers. One feature is specific to innovation. A firm must
survive through the five-year window to contribute, so the estimates describe which firms
go on to innovate rather than the causal effect of improving a workforce's fit.}

\subsection{Mergers and Acquisitions}\label{sec:ma}

Table~\ref{tab:ma} relates how well a target's workers would fit the acquirer, relative
to how well they fit the target, to the share of the purchase price the acquirer records
as goodwill. Goodwill is the natural outcome because of how acquisitions are accounted
for. When one firm buys another, the price it pays above the fair value of identifiable
assets is recorded as goodwill, and part of what goodwill is meant to capture is the
value of the acquired workforce: accounting standards prohibit recognizing an assembled
workforce as a separate intangible asset and fold it into goodwill instead
\citep[ASC 805-20-55-6]{fasb_asc805}, so the human capital an acquirer pays for is,
by construction, an unmeasured residual. \citet{ewens_measuring_2025} use the prices paid
in acquisitions to measure the intangible capital that balance sheets omit. Our question
is narrower and complementary: we focus on one component of what is paid for, the
acquired workforce, and measure it from how well the target's workers fit the acquirer.
Because $\MQ$ is a function of characteristics (Section~\ref{sec:embeddings}), we can ask
not only how well a target's workers fit the target, but how well they would fit the
acquirer.

The distinction matters because the two need not agree. When Facebook acquired
Instagram, what it paid for was not the fit between Instagram's engineers and
Instagram---a firm that would cease to exist as an independent employer---but the fit
between those engineers and Facebook, the organization they were about to join. A
target whose workers are superbly matched to the target as it stands may be one whose
human capital is embedded in that particular organization and travels poorly; a target
whose workers happen to sit close to the acquirer in the embedding space is one whose
human capital the acquirer can use. We construct both objects. \emph{Own} match quality
is the mean of $\MQ_{wft}$ over the target's workers in the year before the deal,
scored against the target's own firm vector. \emph{Counterfactual} match quality takes
the same workers' vectors, computed from their own records and held fixed, and scores
them against the acquirer's firm vector $v_{\text{acq}}$ instead---the same convention
as the candidate ranking of Section~\ref{sec:dest}, applied to a pairing that the deal
is about to realize. Our regressor is the difference between the two, $\Delta\MQ$:
how much better the target's workers would fit the acquirer than they fit the target.

The dependent variable is the log ratio of goodwill to the acquisition price. The
sample starts from 25,500 acquisitions with goodwill reported in XBRL
filings; 9,618 have acquirer and target in the Revelio universe, 4,545 have complete
acquirer controls, 2,369 report an acquisition price, and 1,334 of these have at least
20 scored target employment relationships, so that own and counterfactual match quality
are averaged over enough workers to be meaningful (Panel~D of Appendix
Table~\ref{tab:firmsample}; Appendix Table~\ref{tab:firmdescriptivestats} reports
descriptive statistics).
Standard errors are clustered by acquirer. Column~(1) enters $\Delta\MQ$ alone: the
coefficient is 1.21. Column~(2) adds the acquirer's own pre-deal firm match quality and
the log number of target employment relationships; the coefficient is essentially
unchanged (1.28), and neither addition enters significantly, so the result is not about
acquirers whose own workforces are well matched or about the size of the acquired
workforce. Column~(3)
adds the acquirer's size, profitability, leverage, market-to-book, sales growth,
tangibility, and R\&D intensity, together with indicators for a same-industry deal and a
public target (1.10); column~(4) adds year fixed effects (1.27); and column~(5) adds
acquirer-industry and target-industry fixed effects (0.73, significant at the 5\%
level). Holding fixed the
acquirer's industry and characteristics, the target's industry, and the year, an
acquirer books a larger share of the price as goodwill when the
target's workforce would fit the acquirer better than it fit the target: a 0.1 increase
in $\Delta\MQ$ is associated with a goodwill-to-price ratio about 8\% higher in the
most demanding specification, and about 14\% higher before the industry fixed effects
are added. Appendix
Table~\ref{tab:ma2016} uses the time-split checkpoint and restricts the sample to the
67 deals whose pre-deal match quality is measured in 2016 and whose goodwill is recorded
in 2017; the coefficient is larger there, though with so few deals
it is significant only before the industry fixed effects are added.

Two limits are specific to this setting. Acquirers choose targets, and part of what they
may be choosing on is precisely the fit we measure, so here too the coefficients describe
what is priced rather than the effect of an exogenous change in fit. And goodwill also
contains expected synergies, control premia, and overpayment that have nothing to do
with people; our regressor speaks only to the workforce component. These results rest
on 1,334 deals with at least 20 scored target relationships, a modest sample.
What we would
emphasize is the design rather than the magnitude. The regressor scores a pair that did
not exist when it was computed, which no occupation-level measure or match fixed effect
can do. That the
counterfactual pairing, relative to the realized one, is what the acquirer appears to
pay for is perhaps the most direct evidence in the paper that the measure is doing what
Section~\ref{sec:theory} says it should.

\section{Conclusion}\label{sec:conclusion}

This paper develops a measure of worker--firm match quality that is grounded in
theory, defined at the level of the pair, and computable for any worker and any firm,
whether or not they have been observed together. Match quality is the cosine of worker and firm
embeddings from a two-tower neural network, the inner product of the skill-weights
model \citep{lazear_firmspecific_2009}, and its objective is the separation hazard
implied by the experience-good view of the job match \citep{jovanovic_job_1979}:
durability reveals quality. Because only the timing of
separations enters estimation, every other outcome remains available as a test.
On workers the model has never seen, match quality predicts separations better than a
conventional hazard, and all of that advantage survives as a single term in a logit,
which outperforms the HCT itself. About a quarter of its variance is specific
to the worker--firm pair. Workers earn more when better matched, mainly early in a
relationship, and the match-specific premium fades to zero at long
tenure. The score ranks movers' realized destinations well above chance, and firms
whose workforces score higher grow faster, become more profitable, and over the
following five years are issued more, and more valuable, patents than their industry
peers, and the value association holds within firm as well. The better a target's workers would fit
the acquirer relative to the target, the larger the share of the purchase price the
acquirer records as goodwill. None of these outcomes was a training target.

\paragraph{Interpretation and limitations.} Four qualifications govern how the
results should be read. First, the estimates are associations. The outcomes the
measure predicts are chosen by workers and firms who know things the model does not,
so the coefficients describe what the measure tracks, not the effect of an exogenous
change in fit. What the measure provides is the ability to ask, for the first time at
the level of the pair and at the scale of the economy, how fit relates to pay,
mobility, performance, and the prices paid for workforces; the associations reported
here are a first map of that terrain rather than its causal structure.

Second, the hazard was kept minimal so that other outcomes remain available as tests,
at some cost in predictive efficiency. It contains no
additive worker or firm terms, so $\MQ$ carries worker-level and firm-level separation
propensity that a richer specification would absorb elsewhere, and the pair share of
Table~\ref{tab:pair} is a lower bound. Calendar-year effects are omitted from the
baseline $b(t)$, but absorbing them after the fact leaves the coefficient essentially
unchanged (Table~\ref{tab:hazest}). Auxiliary objectives such as wage growth or
next-employer retrieval could be added and would likely sharpen the score, but they
would consume
the outcomes we use as tests; we kept the objective to separation timing alone so that
everything else could be evidence.

Third, because estimation uses the separation hazard alone, the score cannot by
construction distinguish a durable relationship that is productive from one that is
merely costly to leave. Neither lock-in nor the job ladder implies that
better-matched firms should grow faster, become more profitable, and
patent more, or
that the score should pick out the firms movers join, and both hold in our data. The
wage evidence suggests that lock-in is part of the story for long relationships: the
wage association is larger early in a relationship than after five years at every
level of control, although learning and selection would produce the same gap
(Section~\ref{sec:wages}).

Fourth, the main results use splits by worker rather than by calendar time, so most of
the evidence establishes generalization across people rather than across periods. The
firm-level results, which relate the score to outcomes one to five years ahead, are
already of the temporal form. We also estimate the model with a split in time, training
on data through 2016 only and relating match quality measured in 2016 to outcomes from
2017 onward (Appendix Tables~\ref{tab:firmhor2016}, \ref{tab:innovhor2016},
and~\ref{tab:ma2016}); the results are qualitatively the same. A time-split model with a
longer post-training window is the natural next step as more years of data accrue.

\paragraph{What the embeddings make possible.} The measure is one use of the objects
the model produces. The model delivers a vector for every worker and firm in the same
space, whose geometry is identified even though its coordinates are not
(Section~\ref{sec:embeddings}). Distances between firm vectors define a
human-capital industry classification and peer groups that share a workforce rather
than an industry code; they also measure the human-capital relatedness of an acquirer and
a target, or of a firm and a line of business it is considering, before any
transaction takes place. Worker vectors summarize careers in a form comparable across
people and over time, which bears on skill accumulation, displacement, and the
transferability of experience. And because the score is a function of
characteristics, it can be evaluated for hypothetical pairings---an applicant and a
job, a workforce and a reorganization---which is the form in which human capital
decisions are actually made.

The larger point is about where a measure of human capital can come from. The
measures in use today are either aggregates that treat the workforce as a stock or
firm-level disclosures that describe it in the firm's own words. The measure
developed here is built up from below, one employment relationship at a time, and is
tied to theory both in what it measures and in how it is identified. That it is
associated with what workers are paid, where they go, how their firms perform and
innovate, and what acquirers pay for them suggests that the allocation of people to
firms can be studied with the same granularity as the allocation of capital. For
research, human capital can enter the empirical work of accounting and corporate
finance as something other than a scalar. For practice, a measure of fit defined
before a pairing forms is the input that human capital management has lacked: a way
of asking, for a particular person and a particular firm, whether this is the right
place for that human capital to be.

\newpage

\singlespacing

\bibliography{HCT_biblio}

\clearpage
\section*{Figures and Tables}

\clearpage

\clearpage

\noindent

\begingroup
\centering
\footnotesize
\ShrinkTabulars
\setlength{\parindent}{0pt}
\captionof{table}{Sample Description}
\label{tab:sample}

\begin{tabular}{@{}lrrrr@{}}
\toprule
Variable & N & Mean & SD & Median \\
\midrule
\multicolumn{5}{l}{\textbf{\textit{Panel A: Employment positions}}} \\
\midrule
Employment positions & 340,744,840 &  &  &  \\
Salary & 340,672,446 & 74,231.68 & 54,794.67 & 57,979.89 \\
Seniority & 340,744,840 & 2.37 & 1.54 & 2.00 \\
Role tenure (years) & 251,242,213 & 2.90 & 4.01 & 1.68 \\
Sales share & 340,744,840 & 0.25 &  &  \\
Engineer share & 340,744,840 & 0.20 &  &  \\
Admin share & 340,744,840 & 0.17 &  &  \\
\midrule
\multicolumn{5}{l}{\textbf{\textit{Panel B: Worker--firm relationships}}} \\
\midrule
Worker--firm relationships & 277,448,388 &  &  &  \\
Positions per relationship & 277,448,388 & 1.23 & 0.72 & 1.00 \\
Completed relationship tenure (years) & 194,812,205 & 3.26 & 4.47 & 1.83 \\
Next employer observed at separation & 194,812,205 & 0.87 &  &  \\
Person-period separation rate & 1,176,559,928 & 0.14 &  &  \\
\midrule
\multicolumn{5}{l}{\textbf{\textit{Panel C: Workers}}} \\
\midrule
Workers & 100,647,642 &  &  &  \\
Relationships per worker & 91,529,524 & 3.03 & 2.52 & 2.00 \\
Bachelor's-or-higher share & 100,647,642 & 0.52 &  &  \\
Predicted female share & 100,647,642 & 0.48 &  &  \\
\midrule
\multicolumn{5}{l}{\textbf{\textit{Panel D: Firms}}} \\
\midrule
Firms & 8,384,926 &  &  &  \\
Relationships per firm & 8,384,926 & 33.09 & 1,650.71 & 2.00 \\
``About'' text availability & 8,384,926 & 0.86 &  &  \\
\bottomrule
\end{tabular}

\par\vspace{0.6em}
\begin{minipage}{\textwidth}
\footnotesize This table describes our sample. The hazard is estimated on 277 million
worker--firm relationships observed as 1.18 billion person-periods with a 14\%
separation rate; 87\% of separations have an observed next employer; the
Compustat-linked firms used in Tables~\ref{tab:firm}--\ref{tab:ma} are a subset of a
universe that includes private firms. The sample is the processed Revelio Labs universe
over 2008--2024. A
relationship is the continuous association between a worker and a firm and may comprise
several positions; the annual risk set contains one row (person-period) per relationship per year of
tenure, and
relationships without an observed end date are right-censored and contribute
periods without a separation event. Employment positions are worker--firm--role
observations; for binary variables the mean is a share. Role and relationship tenure are
computed for completed spells only. The Compustat firm-year sample used in
Table~\ref{tab:firm} is described in Appendix Table~\ref{tab:firmsample}; the
innovation sample of Table~\ref{tab:innov} starts from the 71,873 firm-years with at
least 20 scored relationships. Splits are assigned at the worker level (60/20/20).

\end{minipage}
\par
\endgroup
\clearpage

\begin{table}[htbp]
\centering
\footnotesize
\ShrinkTabulars
\caption{Hazard Model Estimates}
\label{tab:hazest}

\begingroup
\centering
\begin{tabular}{lcccc}
   \toprule
   & HCT                             & (1)            & (2)            & (3)\\
   \midrule
   Match Quality (MQ)              & -6.486& -6.457$^{***}$ & -6.490$^{***}$ & -6.510$^{***}$\\
                                   & & (0.0226)       & (0.0256)       & (0.0087)\\
\midrule
   Odds change, +0.1 MQ (\%)       & -47.7& -47.6          & -47.7          & -47.8\\
   Odds change, +1 SD MQ (\%)      & -57.8& -57.6          & -57.8          & -57.9\\
   \midrule
   Tenure fixed effects            & Yes     & Yes            & Yes            & Yes\\
   Calendar-year fixed effects     & &                & Yes            & Yes\\
   Observables, seniority, job cat.& &                &                & Yes\\
\midrule
   Pseudo R$^2$                    & 0.152& 0.160          & 0.164          & 0.165\\
   Workers                         & 4,450,951& 4,450,951      & 4,450,951      & 4,450,945\\
   Firms                           & 1,907,394& 1,907,394      & 1,907,394      & 1,907,375\\
   Observations                    & 58,801,249& 58,801,249     & 58,801,249     & 58,801,178\\
   \bottomrule
\end{tabular}
\par\endgroup

\caption*{\footnotesize This table reports the hazard model estimates. The hazard is
$\lambda_{wft}=\sigma(b(t)+\beta\,\MQ_{wft})$; $\beta$ is
constrained negative as a normalization, since only $\beta\,\MQ_{wft}$ is identified
(Section~\ref{sec:revelation}). The first, unnumbered column is the HCT as estimated
(full-training checkpoint) and reports $\hat\beta$. Columns (1)--(3) are logits of the
separation indicator estimated on the scored person-period file after training, with
$\MQ$ held fixed, not retrained models: column (1) has tenure fixed effects only, column
(2) adds calendar-year fixed effects, and column (3) adds worker and firm observables,
current seniority, and job category.
The logits model the log odds of separation directly, so their coefficients are
comparable with the hazard's $\beta$. Odds changes are for a 0.1 and a
one-standard-deviation increase in $\MQ$, using the standard deviation of $\MQ$ across
person-periods in the estimation sample; the pair-level standard deviation is reported in Table~\ref{tab:pair}. Standard
errors are two-way clustered by worker and firm; $^{***}$ denotes significance at the
1 percent level. Pseudo-$R^2$ is McFadden's. Five-percent analysis subsample of
workers.}

\end{table}
\clearpage

\begin{table}[htbp]
\centering
\footnotesize
\ShrinkTabulars
\caption{Predictive Performance Against Baselines}
\label{tab:hazperf}

\begin{tabular}{@{}lrrrr@{}}
\toprule
 & AUC & BCE & Pseudo-$R^2$ & Person-periods \\
\midrule
\addlinespace[0.5\baselineskip]
\multicolumn{5}{@{}l@{}}{\hspace*{0.5em}\textbf{\textit{Panel A: HCT model}}} \\
\addlinespace[0.5\baselineskip]
\midrule
Tenure baseline & 0.695 & 0.376 & 0.067 & 11,734,024 \\
Full model (Tenure + MQ) & 0.764 & 0.348 & 0.138 & 11,734,024 \\
\midrule
\addlinespace[0.5\baselineskip]
\multicolumn{5}{@{}l@{}}{\hspace*{0.5em}\textbf{\textit{Panel B: Logit models}}} \\
\addlinespace[0.5\baselineskip]
\midrule
Constant separation rate & 0.500 & 0.403 & 0.000 & 11,734,024 \\
Tenure FEs & 0.695 & 0.373 & 0.075 & 11,734,024 \\
Tenure + calendar-year FEs & 0.707 & 0.370 & 0.083 & 11,734,024 \\
Tenure + calendar year + controls & 0.734 & 0.361 & 0.105 & 11,733,975 \\
Tenure + calendar year + controls + MQ & 0.778 & 0.339 & 0.159 & 11,733,975 \\
\bottomrule
\end{tabular}

\caption*{\footnotesize This table reports the predictive performance of the HCT against
conventional baselines on held-out workers. Person-periods is the number
of held-out person-periods evaluated. All rows are fit on training and validation
workers and evaluated on the held-out workers (the 20\% test split) within the
five-percent analysis subsample (training checkpoint). AUC is the area under the
receiver operating characteristic curve for the separation indicator, BCE is binary
cross-entropy, and pseudo-$R^2$ is McFadden's, computed against a constant-rate model on
each row's evaluation sample. Logit rows can use slightly fewer observations when a
fixed-effect category in the held-out data is absent from the training data. Panel~A
reports the HCT with its tenure baseline alone and with $\MQ$; Panel~B reports
logits that add, in turn, tenure, calendar-year, and observable controls, and finally
$\MQ$ as a covariate. The headline comparison is the HCT with $\MQ$ against the logit
with tenure, calendar year, and controls; the last logit row, against the same logit,
shows that the content of $\MQ$ is not an artifact of the neural network that produced
it. Observables are defined in Appendix Table~\ref{tab:vardef}.}

\end{table}
\clearpage

\begin{table}[htbp]
\centering
\footnotesize
\ShrinkTabulars
\caption{Pair-Specific Match Quality}
\label{tab:pair}

\begin{tabular}{@{}lrrrr@{}}
\toprule
& \multicolumn{2}{c}{In-sample} & \multicolumn{2}{c}{Out-of-sample} \\
\cmidrule(lr){2-3} \cmidrule(lr){4-5}
& $R^2$ & Res. SD & $R^2$ & Res. SD \\
\midrule
No FE benchmark & -- & 0.137 & -- & 0.137 \\
Worker FEs & 0.498 & 0.096 & 0.497 & 0.096 \\
Firm FEs & 0.441 & 0.102 & 0.466 & 0.099 \\
Worker + firm FEs & 0.730 & 0.070 & 0.749 & 0.067 \\
\midrule
Number of pairs & \multicolumn{2}{c}{9,494,465} & \multicolumn{2}{c}{2,368,569} \\
Number of workers & \multicolumn{2}{c}{3,562,731} & \multicolumn{2}{c}{888,220} \\
Number of firms & \multicolumn{2}{c}{1,673,351} & \multicolumn{2}{c}{684,169} \\
\bottomrule
\end{tabular}

\caption*{\footnotesize This table reports the variance decomposition of pair-level match
quality. It reports the $R^2$ and residual standard
deviation from regressions of pair-level $\MQ$ on worker fixed effects, firm fixed effects, and both
entered additively; the additive specification contains no worker--firm term, which
would absorb the pair component. Both sets of columns use the training checkpoint on the
five-percent analysis subsample: the in-sample columns score pairs formed by training
and validation workers, and the out-of-sample columns pairs formed by the held-out
workers, whom the checkpoint never saw. The no-fixed-effects row reports the unconditional standard deviation
of $\MQ$, and the pair-specific share is one minus the $R^2$ of the worker-plus-firm row.
Because
the hazard contains no additive worker or firm terms, $\MQ$ must carry worker-level and
firm-level separation propensity, so the pair share is a lower bound. Five-percent
analysis subsample of workers.}
\end{table}

\clearpage

\begin{table}[htbp]
\centering
\footnotesize
\ShrinkTabulars
\caption{Match Quality and Wages}
\label{tab:wage}

\begin{minipage}{\dimexpr\textwidth-4pt\relax}
\centering
\setlength{\tabcolsep}{4pt}
\begingroup
\centering
\begin{tabular*}{\textwidth}{@{\extracolsep{\fill}}lccccc}
   \toprule
                       & (1)           & (2)           & (3)           & (4)           & (5)\\
   \midrule
   \multicolumn{6}{@{}l@{}}{\hspace*{0.5em}\textbf{\textit{Panel A: All relationships}}} \\
   \midrule
   Match Quality (MQ)  & 0.760$^{***}$ & 0.332$^{***}$ & 0.282$^{***}$ & 0.207$^{***}$ & 0.133$^{***}$\\
                       & (0.048)       & (0.005)       & (0.006)       & (0.006)       & (0.005)\\
   \midrule
   R$^2$               & 0.026         & 0.607         & 0.700         & 0.862         & 0.899\\
   Within R$^2$        &               & 0.466         & 0.428         & 0.367         & 0.338\\
   Observations (000s) & 58,799        & 58,799        & 58,528        & 58,617        & 58,361\\
   \bottomrule
\end{tabular*}
\par\endgroup
\par
\par\addvspace{0.25\baselineskip}
\begingroup
\centering
\begin{tabular*}{\textwidth}{@{\extracolsep{\fill}}lccccc}

   \multicolumn{6}{@{}l@{}}{\hspace*{0.5em}\textbf{\textit{Panel B: Tenure 0--4 years}}} \\
   \midrule
   Match Quality (MQ)  & 1.049$^{***}$ & 0.361$^{***}$ & 0.353$^{***}$ & 0.261$^{***}$ & 0.228$^{***}$\\
                       & (0.048)       & (0.006)       & (0.006)       & (0.006)       & (0.006)\\
   \midrule
   R$^2$               & 0.057         & 0.620         & 0.706         & 0.839         & 0.882\\
   Within R$^2$        &               & 0.474         & 0.433         & 0.359         & 0.325\\
   Observations (000s) & 33,555        & 33,555        & 33,270        & 33,334        & 33,067\\
   \bottomrule
\end{tabular*}
\par\endgroup
\par
\par\addvspace{0.25\baselineskip}
\begingroup
\centering
\begin{tabular*}{\textwidth}{@{\extracolsep{\fill}}lccccc}

   \multicolumn{6}{@{}l@{}}{\hspace*{0.5em}\textbf{\textit{Panel C: Tenure 5+ years}}} \\
   \midrule
   Match Quality (MQ)  & 0.129$^{***}$ & 0.280$^{***}$ & 0.186$^{***}$ & 0.122$^{***}$ & 0.007\\
                       & (0.039)       & (0.006)       & (0.010)       & (0.005)       & (0.005)\\
   \midrule
   R$^2$               & 0.001         & 0.573         & 0.700         & 0.930         & 0.958\\
   Within R$^2$        &               & 0.450         & 0.408         & 0.349         & 0.320\\
   Observations (000s) & 25,244        & 25,244        & 25,146        & 24,952        & 24,894\\
   \midrule
   Calendar-year FE    &               & Yes           & Yes           & Yes           & Yes\\
   Tenure-year FE      &               & Yes           & Yes           & Yes           & Yes\\
   Seniority control   &               & Yes           & Yes           & Yes           & Yes\\
   Firm observables    &               & Yes           &               & Yes           & \\
   Worker observables  &               & Yes           & Yes           &               & \\
   Firm FE             &               &               & Yes           &               & Yes\\
   Worker FE           &               &               &               & Yes           & Yes\\
   \bottomrule
\end{tabular*}
\par\endgroup
\par
\end{minipage}

\caption*{\footnotesize This table reports wage regressions on match quality. Cells report
coefficients $\hat{\gamma}$ from regressions of log wage on $\MQ_{wft}$; a 0.1 increase in
match quality corresponds to a wage change of $100(\exp(0.1\hat{\gamma})-1)$ percent, the
units used in the text. Parenthesized values are two-way worker--firm clustered standard
errors; $^{***}$, $^{**}$, $^{*}$ denote significance at the 1, 5, and 10 percent levels.
Observations are in thousands. Worker observables are highest
degree, predicted gender, predicted ethnicity, detailed role category, and location; firm
observables are two-digit industry, coarse and fine Revelio industry classifications, headquarters
metropolitan area and country, and observed size bucket; these are the observables used
in Tables~\ref{tab:hazest} and~\ref{tab:hazperf}. Wages never enter model estimation.
Full-training checkpoint, in sample; five-percent analysis subsample of workers.}
\end{table}

\clearpage

\begin{table}[htbp]
\centering
\footnotesize
\ShrinkTabulars
\caption{Destination Choice at Separations}
\label{tab:dest}

\begin{tabular}{@{}lrrrr@{}}
\toprule
Mover group & Moves & Recall@1 (\%) & Recall@10 (\%) & MRR \\
Random baseline & -- & 0.100 & 1.000 & 0.0075 \\
\midrule
\noalign{\vskip 1\baselineskip}
\multicolumn{5}{@{}l@{}}{\hspace*{0.5em}\textbf{\textit{Panel A: Global candidate pool}}} \\
\addlinespace[0.5\baselineskip]
\midrule
All movers & 9,988 & 0.681 & 3.424 & 0.0193 \\
Same-industry movers & 9,985 & 0.310 & 2.424 & 0.0137 \\
Industry-switching movers & 9,991 & 0.711 & 3.603 & 0.0207 \\
\midrule
\noalign{\vskip 1\baselineskip}
\multicolumn{5}{@{}l@{}}{\hspace*{0.5em}\textbf{\textit{Panel B: Candidates within the destination industry}}} \\
\addlinespace[0.5\baselineskip]
\midrule
All movers & 45,927 & 1.934 & 6.506 & 0.0383 \\
Same-industry movers & 9,508 & 0.652 & 3.807 & 0.0210 \\
Industry-switching movers & 35,822 & 2.286 & 7.244 & 0.0430 \\
\bottomrule
\end{tabular}

\caption*{\footnotesize This table reports how the model ranks a mover's realized destination
among 1,000 candidate firms. The sample is held-out separations with an observed
next employer (training checkpoint). For each event the realized destination is scored
against 999 uniformly sampled, time-valid alternative firms, with the worker vector held
fixed and the firm vector varying. Recall@1 and Recall@10 are the shares of events in
which the destination ranks first and in the top ten; MRR is the mean reciprocal rank.
The random baseline is the expected value under a uniform ordering of the 1,000 firms.
Panel A draws candidates from the global pool. Panel B restricts candidates to the
destination's broad industry; because it takes the industry the worker ends up in as
known, it is not a forecast, and it evaluates ranking only within that industry. Same-industry and
industry-switching rows classify movers by their incumbent and destination broad
industries; the three mover groups in each panel are drawn separately, so their counts
do not sum to the all-movers row.}
\end{table}

\clearpage

\begin{table}[htbp]
\centering
\footnotesize
\ShrinkTabulars
\caption{Firm Performance and Aggregated Match Quality}
\label{tab:firm}

\begin{tabular}{@{}lccc@{}}
\toprule
 & (1) & (2) & (3) \\
\midrule
\multicolumn{4}{@{}l@{}}{\hspace*{0.5em}\textbf{\textit{Panel A. Next-year revenue growth}}} \\
\midrule
Firm MQ & 0.527*** & 0.544*** & 0.704*** \\
 & (0.028) & (0.029) & (0.081) \\
\midrule
R$^2$ & 0.137 & 0.198 & 0.421 \\
\midrule
\multicolumn{4}{@{}l@{}}{\hspace*{0.5em}\textbf{\textit{Panel B. Next-year employment growth}}} \\
\midrule
Firm MQ & 0.406*** & 0.416*** & 0.419*** \\
 & (0.019) & (0.020) & (0.049) \\
\midrule
R$^2$ & 0.157 & 0.193 & 0.395 \\
\midrule
\multicolumn{4}{@{}l@{}}{\hspace*{0.5em}\textbf{\textit{Panel C. Next-year productivity growth}}} \\
\midrule
Firm MQ & 0.072*** & 0.077*** & 0.265*** \\
 & (0.026) & (0.026) & (0.080) \\
\midrule
R$^2$ & 0.083 & 0.132 & 0.352 \\
\midrule
\multicolumn{4}{@{}l@{}}{\hspace*{0.5em}\textbf{\textit{Panel D. Next-year ROA change}}} \\
\midrule
Firm MQ & 0.137*** & 0.142*** & 0.045 \\
 & (0.013) & (0.013) & (0.038) \\
\midrule
R$^2$ & 0.192 & 0.233 & 0.450 \\
\midrule
Controls & Yes & Yes & Yes \\
Industry FE & Yes &  &  \\
Year FE & Yes &  &  \\
Industry $\times$ year FE &  & Yes & Yes \\
Firm FE &  &  & Yes \\
\midrule
Observations & 48,336 & 48,336 & 47,775 \\
\bottomrule
\end{tabular}

\caption*{\footnotesize This table reports regressions of next-year firm outcomes on firm
match quality. Panels A--C use next-year log revenue, employment, and
revenue-per-employee growth; Panel D uses $ROA_{t+1}-ROA_t$. Firm MQ (firm match quality) is the firm-year
mean of $\MQ_{wft}$ over scored employment relationships, in original units. All columns control for the outcome's level and lagged change, log scored relationships,
log current employees, R\&D intensity and an R\&D-missing indicator, log assets, log
market value, book-to-market, leverage, tangibility, an M\&A indicator (with a missing
indicator), return volatility, and the standard deviation of $\MQ_{wft}$ across the
firm's scored relationships,
together with the indicated fixed effects; industries
are three-digit NAICS, and singleton industry-by-year cells are dropped. Estimates over
horizons of two to five years are in Appendix
Tables~\ref{tab:firmhor}--\ref{tab:firmhorfe}. Standard errors are clustered by firm.
No financial variable enters model estimation. Compustat firms with at least 20 scored
relationships, 2008--2024 (Appendix Table~\ref{tab:firmsample}). $^{***}$, $^{**}$,
$^{*}$: significance at 1, 5, and 10 percent.}
\end{table}

\clearpage

\begin{table}[htbp]
\centering
\footnotesize
\ShrinkTabulars
\caption{Innovation and Aggregated Match Quality}
\label{tab:innov}

\begin{tabular}{lccc}
\toprule
 & (1) & (2) & (3) \\
\midrule
\multicolumn{4}{@{}l@{}}{\hspace*{0.5em}\textbf{\textit{Panel A. Cumulative patents issued, t+1 to t+5}}} \\
\midrule
Firm MQ & 1.448*** & 1.453*** & 1.411* \\
 & (0.373) & (0.378) & (0.742) \\
Firm MQ SD & -0.976 & 0.548 & -1.877 \\
 & (1.905) & (1.977) & (1.841) \\
\midrule
Pseudo R$^2$ & 0.916 & 0.921 & 0.972 \\
Observations & 35,944 & 35,258 & 20,545 \\
\midrule
\multicolumn{4}{@{}l@{}}{\hspace*{0.5em}\textbf{\textit{Panel B. Citations per issued patent, t+1 to t+5}}} \\
\midrule
Firm MQ & 0.422* & 0.428* & 0.077 \\
 & (0.219) & (0.225) & (0.368) \\
Firm MQ SD & 2.103** & 2.262** & 0.685 \\
 & (0.987) & (1.020) & (0.719) \\
\midrule
R$^2$ & 0.388 & 0.399 & 0.853 \\
Observations & 17,716 & 17,555 & 17,264 \\
\midrule
\multicolumn{4}{@{}l@{}}{\hspace*{0.5em}\textbf{\textit{Panel C. Real value per issued patent, t+1 to t+5}}} \\
\midrule
Firm MQ & 1.687*** & 1.692*** & 1.713*** \\
 & (0.246) & (0.252) & (0.400) \\
Firm MQ SD & 2.540* & 2.685* & 1.751** \\
 & (1.325) & (1.382) & (0.681) \\
\midrule
R$^2$ & 0.737 & 0.742 & 0.961 \\
Observations & 17,716 & 17,555 & 17,264 \\
\midrule
Controls & Yes & Yes & Yes \\
Industry FE & Yes &  &  \\
Year FE & Yes &  &  \\
Industry $\times$ year FE &  & Yes & Yes \\
Firm FE &  &  & Yes \\
\bottomrule
\end{tabular}

\caption*{\footnotesize This table reports regressions of innovation outcomes on firm match
quality. Outcomes are cumulated from $t+1$ through $t+5$. Panel A reports the
count of patents issued, estimated by zero-inclusive Poisson pseudo-maximum likelihood.
Panel B reports $\log(1+\text{citations}/\text{patents})$ for issued patents,
conditional on a positive patent count; Panel C reports the log ratio of real patent
value to issued patents, conditional on a positive patent count and positive value;
both outcomes are winsorized at the first and ninety-ninth percentiles and estimated by
OLS.
Patent data are from \citet{kogan_technological_2017}, with 2024 the last usable patent
year, so match quality is measured through 2019. Firm MQ is in original units; Firm MQ
SD is the standard deviation of $\MQ_{wft}$ across the firm's scored relationships;
continuous controls are winsorized at the first and ninety-ninth percentiles. All
columns include the controls of Table~\ref{tab:firm}; column (1) adds industry and
year fixed effects, column (2) three-digit NAICS industry-by-year fixed effects, and
column (3) firm fixed effects as well. Cumulative horizons from $t+1$ to $t+5$ are in
Appendix Tables~\ref{tab:innovhor}--\ref{tab:innovhorfe}. Standard errors are
clustered by firm. No patent variable enters model estimation. $^{***}$, $^{**}$,
$^{*}$: significance at 1, 5, and 10 percent.}

\end{table}
\clearpage

\begin{table}[htbp]
\centering
\footnotesize
\ShrinkTabulars
\caption{Acquisitions: Counterfactual Match Quality of the Acquired Workforce}
\label{tab:ma}

\begingroup
\centering
\begin{tabular}{lccccc}
   \toprule
                                & (1)             & (2)             & (3)             & (4)             & (5)\\
   \midrule
   Delta MQ                     & 1.207$^{***}$   & 1.284$^{***}$   & 1.097$^{***}$   & 1.274$^{***}$   & 0.7321$^{**}$\\
                                & (0.3107)        & (0.3274)        & (0.3183)        & (0.3590)        & (0.3384)\\
   Pre-deal acquirer MQ         &                 & -0.2219         & -0.3775         & -0.6611         & -0.2214\\
                                &                 & (0.5180)        & (0.5018)        & (0.5349)        & (0.5015)\\
   Log target relationships     &                 & 0.0034          & 0.0389          & 0.0338          & 0.0032\\
                                &                 & (0.0188)        & (0.0240)        & (0.0242)        & (0.0242)\\
   Log acquirer assets          &                 &                 & 0.0325$^{**}$   & 0.0342$^{**}$   & 0.0274$^{*}$\\
                                &                 &                 & (0.0151)        & (0.0156)        & (0.0140)\\
   Acquirer ROA                 &                 &                 & -0.4401$^{***}$ & -0.4276$^{***}$ & -0.2988$^{*}$\\
                                &                 &                 & (0.1576)        & (0.1591)        & (0.1584)\\
   Acquirer leverage            &                 &                 & -0.2655$^{**}$  & -0.2886$^{**}$  & -0.2472$^{**}$\\
                                &                 &                 & (0.1310)        & (0.1326)        & (0.1250)\\
   Acquirer M/B                 &                 &                 & 0.0122$^{***}$  & 0.0113$^{***}$  & 0.0090$^{***}$\\
                                &                 &                 & (0.0023)        & (0.0023)        & (0.0022)\\
   Acquirer sales growth        &                 &                 & 0.0559          & 0.0467          & -0.0088\\
                                &                 &                 & (0.0587)        & (0.0582)        & (0.0545)\\
   Acquirer tangibility         &                 &                 & -0.6223$^{***}$ & -0.5817$^{***}$ & -0.1252\\
                                &                 &                 & (0.1712)        & (0.1728)        & (0.1664)\\
   Acquirer R\&D/assets         &                 &                 & -0.1024         & -0.1871         & -0.4669\\
                                &                 &                 & (0.4467)        & (0.4584)        & (0.4472)\\
   R\&D missing                 &                 &                 & 0.0324          & 0.0261          & -0.0773\\
                                &                 &                 & (0.0557)        & (0.0548)        & (0.0513)\\
   Same industry                &                 &                 & -0.0236         & -0.0276         & -0.0439\\
                                &                 &                 & (0.0415)        & (0.0411)        & (0.0394)\\
   Public target                &                 &                 & -0.2500$^{***}$ & -0.2534$^{***}$ & -0.1584$^{**}$\\
                                &                 &                 & (0.0765)        & (0.0787)        & (0.0740)\\
   Constant                     & -0.8387$^{***}$ & -0.8539$^{***}$ & -1.136$^{***}$  &                 &   \\
                                & (0.0269)        & (0.0880)        & (0.1344)        &                 &   \\
   \midrule
   Observations                 & 1,334           & 1,334           & 1,334           & 1,334           & 1,334\\
   R$^2$                        & 0.01615         & 0.01640         & 0.08321         & 0.10507         & 0.19734\\
   \midrule
   Year fixed effects           &                 &                 &                 & Yes             & Yes\\
   Acquirer NAICS fixed effects &                 &                 &                 &                 & Yes\\
   Target NAICS fixed effects   &                 &                 &                 &                 & Yes\\
   \bottomrule
\end{tabular}
\par\endgroup

\caption*{\footnotesize This table reports regressions of the goodwill share of the purchase
price on counterfactual match quality. The dependent variable is the log ratio of goodwill to acquisition price, winsorized at
the first and ninety-ninth percentiles. $\Delta\MQ$ is counterfactual match quality, the
score of each target worker's pre-deal embedding against the acquirer's firm vector,
averaged over target workers, minus the same workers' own match quality at the target.
Columns (2)--(5) add pre-deal acquirer match quality and log target employment
relationships; columns (3)--(5) add acquirer log assets, return on assets, leverage,
market-to-book, sales growth, tangibility, R\&D to assets with an R\&D-missing indicator,
and indicators for a same-industry deal and a public target; columns (4)--(5) add
year fixed effects, and column (5) adds acquirer-industry and target-industry fixed
effects (one-digit NAICS sectors). Deals are acquisitions with goodwill reported in XBRL filings, with acquirer
and target in the Revelio universe, complete controls, an acquisition price, and at
least 20 scored target employment relationships (1,334 deals; Panel~D of Appendix
Table~\ref{tab:firmsample}). All columns use the same deals. Standard
errors are clustered by acquirer. $^{***}$, $^{**}$, $^{*}$: significance at 1, 5, and
10 percent.}
\end{table}

\clearpage

\appendix
\section*{Appendix Tables}
\setcounter{table}{0}
\renewcommand{\thetable}{A.\arabic{table}}
\renewcommand{\theHtable}{A.\arabic{table}}

\begingroup
\footnotesize
\setlength{\LTcapwidth}{\textwidth}
\begin{longtable}{@{}>{\raggedright\arraybackslash}p{0.27\textwidth}>{\raggedright\arraybackslash}p{0.70\textwidth}@{}}
\caption{Variable Definitions}\label{tab:vardef}\\
\toprule

Variable & Definition \\
\midrule
\endfirsthead
\multicolumn{2}{@{}l}{\footnotesize\textit{Table~\thetable, continued}}\\
\toprule
Variable & Definition \\
\midrule
\endhead
\midrule
\multicolumn{2}{r@{}}{\footnotesize\textit{continued on next page}}\\
\endfoot
\bottomrule
\endlastfoot
\multicolumn{2}{@{}l@{}}{\hspace*{0.5em}\textbf{\textit{Panel A. Revelio Labs and model-derived employment measures}}} \\
\midrule
Employment relationship & The continuous association between a worker and a firm, comprising every position the worker holds at that firm. Reconstructed from Revelio position-level events. \\
Person-period & One row per relationship per year of tenure. \\
Separation indicator ($d_{wft}$) & Equals one if the relationship ends during the person-period. \\
Match quality ($\MQ_{wft}$) & Weighted cosine similarity between the worker and firm embeddings (equation~\eqref{eq:mq}); bounded in $[-1,1]$. \\
Tenure ($t$) & Years since the start of the relationship. \\
Worker observables & Highest degree, predicted gender, predicted ethnicity, detailed role category, and location. \\
Firm observables & Two-digit industry, coarse and fine Revelio industry classifications, headquarters metropolitan area, headquarters country, and observed size bucket. \\
Current seniority, job category & The seniority level (an integer scale) and job category of the position the worker holds in the current person-period. \\
Log wage & Log of the Revelio salary estimate for the position held in the person-period. Salaries are modeled by Revelio from position characteristics, not observed. \\
Firm MQ (firm match quality) & Mean of the model's match-quality scores over the firm's scored employment relationships in the year. \\
Firm MQ SD & Standard deviation of the model's match-quality scores across the firm's scored relationships in the year. \\
Log scored relationships & Log of the number of the firm's employment relationships scored by the model in the year. \\
Recall@1, Recall@10, MRR & Share of moves whose realized destination ranks first (in the top ten) among 1,000 candidate firms; MRR is the mean reciprocal rank of the destination (Table~\ref{tab:dest}). \\
\addlinespace
\midrule
\multicolumn{2}{@{}l@{}}{\hspace*{0.5em}\textbf{\textit{Panel B. Compustat and CRSP firm-year measures}}} \\
\midrule
Return on assets (ROA) & Net income divided by total assets (\texttt{ni}/\texttt{at}). \\
ROA change & $ROA_{t+1}-ROA_t$; at longer horizons, average ROA over $t+1$ through $t+h$ minus $ROA_t$. \\
Revenue, employment, and productivity growth & Log changes in Compustat revenue (\texttt{revt}), employees (\texttt{emp}), and revenue per employee. \\
Log current employees & Log of Compustat employees (\texttt{emp}) in year $t$. \\
Outcome baseline & The level of each outcome in year $t$ and its change from $t-1$ to $t$, included as controls in Table~\ref{tab:firm}. \\
R\&D intensity & R\&D expense (\texttt{xrd}) divided by total assets (\texttt{at}). Set to zero when missing, with a separate missing-R\&D indicator. \\
Log market value & Log of equity market value, calculated as common shares outstanding times fiscal-year-end price (\texttt{csho}$\times$\texttt{prcc\_f}). \\
Book-to-market & Book equity divided by equity market value (\texttt{ceq}/(\texttt{csho}$\times$\texttt{prcc\_f})). \\
Leverage & Current debt plus long-term debt divided by total assets ((\texttt{dlc}+\texttt{dltt})/\texttt{at}). Set to zero when missing. \\
Asset tangibility & Property, plant, and equipment (\texttt{ppent}) divided by total assets (\texttt{at}). \\
M\&A indicator & Equals one when Compustat acquisitions (\texttt{aqc}) are positive in the year. A separate indicator equals one when \texttt{aqc} is missing. \\
Log assets & Log of total assets (\texttt{at}). \\
Return volatility & Annualized standard deviation of monthly CRSP returns over the trailing 12 months. Requires at least six nonmissing monthly returns. \\
\addlinespace
\midrule
\multicolumn{2}{@{}l@{}}{\hspace*{0.5em}\textbf{\textit{Panel C. Patent data from \citet{kogan_technological_2017}}}} \\
\midrule
Patents issued & Number of patents issued to the firm from $t+1$ through $t+n$. \\
Citations per patent & $\log(1+\text{citations}/\text{patents})$ for the firm's issued patents over the window. \\
Real value per patent & Log of the ratio of the real market value of the firm's issued patents over the window to their number. \\
\addlinespace
\midrule
\multicolumn{2}{@{}l@{}}{\hspace*{0.5em}\textbf{\textit{Panel D. XBRL acquisitions and workforce measures}}} \\
\midrule
Goodwill & Goodwill recognized in the acquisition. Selected from the XBRL concepts \texttt{Goodwill}, \texttt{BusinessAcquisitionPurchasePriceAllocationGoodwillAmount}, \texttt{BusinessAcquisitionsPurchasePriceAllocationGoodwillAmount}, and \texttt{GoodwillGross}, in that order of preference. \\
Acquisition (purchase) price & Total consideration transferred for the business combination, from XBRL concept \texttt{BusinessCombinationConsiderationTransferred1}. \\
Log goodwill-to-price ratio & Natural logarithm of goodwill divided by purchase price. \\
Own $\MQ$ of target workers & Mean of $\MQ_{wft}$ over the target's workers in the year before the deal, scored against the target's own firm vector. \\
Counterfactual $\MQ$ of target workers & Mean over the same workers of the weighted cosine, as in $\MQ_{wft}$, between each worker's vector, computed from the worker's own record and held fixed, and the acquirer's firm vector. \\
$\Delta\MQ$ & Counterfactual minus own $\MQ$ of the target's workers (Table~\ref{tab:ma}). \\
Pre-deal acquirer MQ & Firm MQ of the acquirer in the year before the deal. \\
Log target relationships & Log of the number of the target's scored employment relationships in the year before the deal. \\
Acquirer controls & Log assets, ROA, leverage, market-to-book (equity market value divided by book equity), sales growth (log change in revenue), tangibility, and R\&D divided by assets with an R\&D-missing indicator, each measured for the acquirer in the fiscal year before the deal; and indicators for a deal in which acquirer and target share an industry classification and for a publicly traded target. \\

\end{longtable}
\vspace{-1.2em}
\noindent\parbox{\textwidth}{\footnotesize Firm-year regression variables---outcomes, controls, Firm MQ, and Firm MQ SD---are winsorized at the first and ninety-ninth percentiles. In the XBRL analysis, the log goodwill-to-price ratio is winsorized at the same percentiles. Firm-year regressions use three-digit NAICS industries. Compustat/CRSP, patent, XBRL, wage, and destination variables do not enter model estimation.}
\endgroup

\clearpage

\begin{table}[htbp]
\centering
\footnotesize
\ShrinkTabulars
\caption{Firm-Level Sample Construction}
\label{tab:firmsample}

\begin{tabular}{@{}lrr@{}}
\toprule
Restriction & Firm-years & Firms \\
\midrule
\multicolumn{3}{l}{\textbf{\textit{Panel A: Common firm-year sample}}} \\
\midrule
Compustat firm-years & 163,331 & 20,654 \\
with MQ score (Revelio) & 94,459 & 10,064 \\
with at least 20 MQ scores & 71,873 & 7,705 \\
\midrule
\multicolumn{3}{l}{\textbf{\textit{Panel B: Firm performance}}} \\
\midrule
with available controls and outcomes & 48,432 & 5,623 \\
with non-singleton industry-year cells & 48,336 & 5,622 \\
\midrule
\multicolumn{3}{l}{\textbf{\textit{Panel C: Innovation}}} \\
\midrule
with available controls & 51,606 & 5,987 \\
Controls and t+5 patents & 35,994 & 4,994 \\
Controls and t+5 citations, positive patents & 17,717 & 2,660 \\
Controls and t+5 value, positive patents and value & 17,717 & 2,660 \\
\midrule
\multicolumn{3}{l}{\textbf{\textit{Panel D: M\&A}}} \\
\midrule
Restriction & Deals & Acquirers \\
\midrule
With goodwill (MQ year through 2024) & 25,500 & 4,768 \\
With acquirer and target in Revelio & 9,618 & 3,238 \\
With available controls & 4,545 & 1,783 \\
With acquisition price & 2,369 & 1,194 \\
With at least 20 target MQ scores & 1,334 & 815 \\
\bottomrule
\end{tabular}

\caption*{\footnotesize This table reports the construction of the firm-level samples. Panel~A starts from all Compustat firm-years over 2008--2024, retains approved one-to-one Revelio--Compustat links with a finite firm match quality, and then requires at least 20 scored employment relationships. Panel~B requires all four next-year performance outcomes and the controls of Table~\ref{tab:firm}, and then drops singleton three-digit NAICS industry-by-year cells; the last row is the sample of Table~\ref{tab:firm}. Panel~C requires the controls of Table~\ref{tab:innov} and then the five-year cumulative outcomes: issued patents for all firm-years with patent data, and citations and real value per patent for firm-years with at least one issued patent. Panel~D reports deal counts for the acquisition sample of Table~\ref{tab:ma} after sequential restrictions: goodwill reported in XBRL filings with pre-deal match quality measured through 2024, acquirer and target in the Revelio universe, available controls (allowing missing R\&D, since the specification includes an R\&D-missing indicator), an acquisition price, and at least 20 scored target employment relationships.}
\end{table}
\clearpage

\begin{table}[htbp]
\centering
\footnotesize
\ShrinkTabulars
\caption{Descriptive Statistics for the Firm-Level Samples}
\label{tab:firmdescriptivestats}

\begin{tabular}{@{}lrrrrrr@{}}
\toprule
Variable & N & Mean & SD & P25 & Median & P75 \\
\midrule
\multicolumn{7}{l}{\textbf{\textit{Panel A: Firm-year characteristics}}} \\
\midrule
Firm MQ & 48,336 & 0.02 & 0.07 & -0.03 & 0.01 & 0.06 \\
Firm MQ SD & 48,336 & 0.10 & 0.02 & 0.09 & 0.09 & 0.11 \\
Log relationships & 48,336 & 6.34 & 1.82 & 4.93 & 6.30 & 7.61 \\
Log employees & 48,336 & 0.82 & 2.12 & -0.74 & 0.87 & 2.35 \\
ROA & 48,336 & -0.03 & 0.23 & -0.04 & 0.02 & 0.07 \\
R\&D intensity & 48,336 & 0.05 & 0.11 & 0.00 & 0.00 & 0.06 \\
Log assets & 48,336 & 7.24 & 2.22 & 5.70 & 7.22 & 8.73 \\
Log market value & 48,336 & 7.09 & 2.19 & 5.57 & 7.12 & 8.59 \\
Book-to-market & 48,336 & 0.56 & 0.58 & 0.22 & 0.45 & 0.78 \\
Leverage & 48,336 & 0.25 & 0.23 & 0.06 & 0.21 & 0.38 \\
Tangibility & 48,336 & 0.22 & 0.23 & 0.05 & 0.13 & 0.33 \\
Return volatility & 48,336 & 0.46 & 0.29 & 0.26 & 0.38 & 0.56 \\
M\&A indicator & 48,336 & 0.37 & 0.48 &  & 0.00 &  \\
\midrule
\multicolumn{7}{l}{\textbf{\textit{Panel B: Firm performance outcomes}}} \\
\midrule
Next-year log revenue growth & 48,336 & 0.05 & 0.31 & -0.04 & 0.05 & 0.15 \\
Next-year log employment growth & 48,336 & 0.02 & 0.20 & -0.04 & 0.02 & 0.09 \\
Next-year log revenue-per-employee growth & 48,336 & 0.03 & 0.29 & -0.06 & 0.02 & 0.11 \\
Change in ROA through t+1 & 48,336 & 0.00 & 0.16 & -0.03 & 0.00 & 0.03 \\
\midrule
\multicolumn{7}{l}{\textbf{\textit{Panel C: Innovation outcomes}}} \\
\midrule
Issued patents (t+1 to t+5) & 35,994 & 108.12 & 463.87 & 0.00 & 0.00 & 16.00 \\
Log(1 + citations per issued patent), t+1 to t+5 & 17,717 & 1.55 & 0.97 & 0.83 & 1.47 & 2.16 \\
Log real value per issued patent, t+1 to t+5 & 17,717 & 1.37 & 1.67 & 0.24 & 1.44 & 2.50 \\
\midrule
\multicolumn{7}{l}{\textbf{\textit{Panel D: M\&A}}} \\
\midrule
Log goodwill/acquisition price & 1,334 & -0.80 & 0.74 & -1.07 & -0.62 & -0.35 \\
Delta MQ & 1,334 & 0.04 & 0.08 & -0.01 & 0.03 & 0.08 \\
Pre-deal acquirer MQ & 1,334 & 0.02 & 0.05 & -0.02 & 0.02 & 0.05 \\
Log target relationships & 1,334 & 4.58 & 1.24 & 3.64 & 4.28 & 5.20 \\
Log acquirer assets & 1,334 & 7.67 & 1.78 & 6.50 & 7.67 & 8.79 \\
Acquirer ROA & 1,334 & 0.01 & 0.13 & -0.01 & 0.03 & 0.07 \\
Acquirer leverage & 1,334 & 0.25 & 0.18 & 0.10 & 0.24 & 0.37 \\
Acquirer market-to-book & 1,334 & 5.85 & 7.64 & 2.01 & 3.41 & 6.16 \\
Acquirer sales growth & 1,334 & 0.18 & 0.32 & 0.03 & 0.11 & 0.25 \\
Acquirer tangibility & 1,334 & 0.14 & 0.15 & 0.04 & 0.09 & 0.18 \\
Acquirer R\&D/assets & 1,334 & 0.04 & 0.06 & 0.00 & 0.01 & 0.06 \\
Same industry & 1,334 & 0.56 & 0.50 &  & 1.00 &  \\
Public target & 1,334 & 0.14 & 0.35 &  & 0.00 &  \\
\bottomrule
\end{tabular}

\caption*{\footnotesize This table reports descriptive statistics for the variables of the firm-level analyses. Panels~A and~B describe the firm-year sample of Table~\ref{tab:firm} (48,336 firm-years). Panel~C describes the innovation outcomes of Table~\ref{tab:innov}, cumulated from $t+1$ through $t+5$, for the firm-years eligible for its regressions (Panel~C of Appendix Table~\ref{tab:firmsample}), before fixed-effect singletons are dropped. Panel~D describes the deal sample of Table~\ref{tab:ma} (1,334 deals). Continuous firm-year variables are winsorized at the first and ninety-ninth percentiles. Variables are defined in Appendix Table~\ref{tab:vardef}.}
\end{table}
\clearpage

\begin{table}[htbp]
\centering
\footnotesize
\ShrinkTabulars
\caption{Firm Performance over Future Horizons: Industry $\times$ Year Fixed Effects}
\label{tab:firmhor}

\begin{tabular}{lccccc}
\toprule
 & $t+1$ & $t+2$ & $t+3$ & $t+4$ & $t+5$ \\
\midrule
\multicolumn{6}{@{}l@{}}{\hspace*{0.5em}\textbf{\textit{Panel A. Revenue growth}}} \\
\midrule
Firm MQ & 0.544*** & 0.562*** & 0.558*** & 0.536*** & 0.502*** \\
 & (0.029) & (0.028) & (0.027) & (0.026) & (0.026) \\
\midrule
R$^2$ & 0.198 & 0.228 & 0.245 & 0.260 & 0.276 \\
\midrule
\multicolumn{6}{@{}l@{}}{\hspace*{0.5em}\textbf{\textit{Panel B. Employment growth}}} \\
\midrule
Firm MQ & 0.416*** & 0.452*** & 0.462*** & 0.455*** & 0.429*** \\
 & (0.020) & (0.021) & (0.022) & (0.022) & (0.023) \\
\midrule
R$^2$ & 0.193 & 0.205 & 0.213 & 0.221 & 0.233 \\
\midrule
\multicolumn{6}{@{}l@{}}{\hspace*{0.5em}\textbf{\textit{Panel C. Productivity growth}}} \\
\midrule
Firm MQ & 0.077*** & 0.068*** & 0.063*** & 0.061*** & 0.060*** \\
 & (0.026) & (0.022) & (0.020) & (0.019) & (0.017) \\
\midrule
R$^2$ & 0.132 & 0.171 & 0.205 & 0.231 & 0.248 \\
\midrule
\multicolumn{6}{@{}l@{}}{\hspace*{0.5em}\textbf{\textit{Panel D. ROA change}}} \\
\midrule
Firm MQ & 0.142*** & 0.175*** & 0.189*** & 0.197*** & 0.195*** \\
 & (0.013) & (0.016) & (0.017) & (0.018) & (0.019) \\
\midrule
R$^2$ & 0.233 & 0.304 & 0.357 & 0.394 & 0.440 \\
\midrule
Observations & 48,336 & 43,111 & 38,216 & 33,696 & 29,643 \\
\bottomrule
\end{tabular}

\caption*{\footnotesize This table extends Table~\ref{tab:firm} to horizons of two to five years with industry-by-year fixed effects. Each specification includes the controls of Table~\ref{tab:firm}, outcome baselines, and three-digit NAICS industry-by-year fixed effects; the $t+1$ column repeats column (2) of Table~\ref{tab:firm}. Revenue, employment, and productivity outcomes are annualized cumulative log changes from $t$ through the indicated horizon. ROA change is average ROA over $t+1$ through the horizon minus $ROA_t$. Standard errors are clustered by firm. $^{***}$, $^{**}$, $^{*}$: significance at 1, 5, and 10 percent.}
\end{table}
\clearpage

\begin{table}[htbp]
\centering
\footnotesize
\ShrinkTabulars
\caption{Firm Performance over Future Horizons: Firm and Industry $\times$ Year Fixed Effects}
\label{tab:firmhorfe}

\begin{tabular}{lccccc}
\toprule
 & $t+1$ & $t+2$ & $t+3$ & $t+4$ & $t+5$ \\
\midrule
\multicolumn{6}{@{}l@{}}{\hspace*{0.5em}\textbf{\textit{Panel A. Revenue growth}}} \\
\midrule
Firm MQ & 0.704*** & 0.587*** & 0.536*** & 0.486*** & 0.464*** \\
 & (0.081) & (0.074) & (0.071) & (0.068) & (0.065) \\
\midrule
R$^2$ & 0.421 & 0.544 & 0.628 & 0.691 & 0.748 \\
\midrule
\multicolumn{6}{@{}l@{}}{\hspace*{0.5em}\textbf{\textit{Panel B. Employment growth}}} \\
\midrule
Firm MQ & 0.419*** & 0.373*** & 0.371*** & 0.350*** & 0.312*** \\
 & (0.049) & (0.050) & (0.050) & (0.048) & (0.045) \\
\midrule
R$^2$ & 0.395 & 0.529 & 0.619 & 0.695 & 0.758 \\
\midrule
\multicolumn{6}{@{}l@{}}{\hspace*{0.5em}\textbf{\textit{Panel C. Productivity growth}}} \\
\midrule
Firm MQ & 0.265*** & 0.180*** & 0.142*** & 0.119** & 0.116*** \\
 & (0.080) & (0.064) & (0.054) & (0.049) & (0.044) \\
\midrule
R$^2$ & 0.352 & 0.497 & 0.587 & 0.651 & 0.706 \\
\midrule
\multicolumn{6}{@{}l@{}}{\hspace*{0.5em}\textbf{\textit{Panel D. ROA change}}} \\
\midrule
Firm MQ & 0.045 & 0.115*** & 0.191*** & 0.234*** & 0.257*** \\
 & (0.038) & (0.043) & (0.048) & (0.049) & (0.050) \\
\midrule
R$^2$ & 0.450 & 0.610 & 0.691 & 0.742 & 0.785 \\
\midrule
Observations & 47,775 & 42,591 & 37,618 & 33,166 & 29,257 \\
\bottomrule
\end{tabular}

\caption*{\footnotesize This table extends Table~\ref{tab:firm} to horizons of two to five years with firm and industry-by-year fixed effects. It uses the outcomes, controls, and horizon definitions of Appendix Table~\ref{tab:firmhor}, adding firm fixed effects to the industry-by-year fixed effects; the $t+1$ column repeats column (3) of Table~\ref{tab:firm}. Standard errors are clustered by firm. $^{***}$, $^{**}$, $^{*}$: significance at 1, 5, and 10 percent.}
\end{table}
\clearpage

\begin{table}[htbp]
\centering
\footnotesize
\ShrinkTabulars
\caption{Firm Performance with Match Quality Fixed in 2016}
\label{tab:firmhor2016}

\begin{tabular}{lccccc}
\toprule
 & $t+1$ (2017) & $t+2$ (2018) & $t+3$ (2019) & $t+4$ (2020) & $t+5$ (2021) \\
\midrule
\multicolumn{6}{@{}l@{}}{\hspace*{0.5em}\textbf{\textit{Revenue growth}}} \\
\midrule
Firm MQ & 0.412*** & 0.394*** & 0.270*** & 0.255*** & 0.208*** \\
 & (0.086) & (0.070) & (0.054) & (0.048) & (0.047) \\
\midrule
R$^2$ & 0.138 & 0.191 & 0.175 & 0.284 & 0.245 \\
Observations & 2,768 & 2,613 & 2,479 & 2,360 & 2,239 \\
\midrule
\multicolumn{6}{@{}l@{}}{\hspace*{0.5em}\textbf{\textit{Employment growth}}} \\
\midrule
Firm MQ & 0.319*** & 0.275*** & 0.199*** & 0.220*** & 0.161*** \\
 & (0.058) & (0.048) & (0.040) & (0.038) & (0.034) \\
\midrule
R$^2$ & 0.163 & 0.174 & 0.188 & 0.208 & 0.211 \\
Observations & 2,768 & 2,613 & 2,479 & 2,360 & 2,239 \\
\midrule
\multicolumn{6}{@{}l@{}}{\hspace*{0.5em}\textbf{\textit{Productivity growth}}} \\
\midrule
Firm MQ & 0.013 & 0.082 & 0.051 & 0.043 & 0.032 \\
 & (0.087) & (0.060) & (0.044) & (0.038) & (0.036) \\
\midrule
R$^2$ & 0.100 & 0.158 & 0.154 & 0.265 & 0.238 \\
Observations & 2,768 & 2,613 & 2,479 & 2,360 & 2,239 \\
\midrule
\multicolumn{6}{@{}l@{}}{\hspace*{0.5em}\textbf{\textit{ROA change}}} \\
\midrule
Firm MQ & 0.038 & 0.005 & 0.010 & 0.004 & -0.041 \\
 & (0.043) & (0.044) & (0.040) & (0.035) & (0.035) \\
\midrule
R$^2$ & 0.174 & 0.212 & 0.254 & 0.342 & 0.408 \\
Observations & 2,768 & 2,613 & 2,479 & 2,360 & 2,239 \\
\midrule
Controls & Yes & Yes & Yes & Yes & Yes \\
Industry FE & Yes & Yes & Yes & Yes & Yes \\
\bottomrule
\end{tabular}

\caption*{\footnotesize This table repeats the cross-firm specification of Table~\ref{tab:firm} with firm match quality scored by the time-split checkpoint, which is trained on data through 2016 only. Match quality and all controls are measured in a single year, 2016, and the outcomes at horizons of one to five years, 2017 through 2021, so that no outcome overlaps the years the model was trained on. Outcomes are defined as in Appendix Table~\ref{tab:firmhor} over the window from 2016 through the indicated year. Each specification includes the controls of Table~\ref{tab:firm} and three-digit NAICS industry fixed effects; firm and industry-by-year fixed effects cannot be estimated in a single cross-section. Standard errors are clustered by firm. $^{***}$, $^{**}$, $^{*}$: significance at 1, 5, and 10 percent.}
\end{table}
\clearpage

\begin{table}[htbp]
\centering
\footnotesize
\ShrinkTabulars
\caption{Innovation over Future Horizons: Industry $\times$ Year Fixed Effects}
\label{tab:innovhor}

\begin{tabular}{lccccc}
\toprule
 & $t+1$ & $t+2$ & $t+3$ & $t+4$ & $t+5$ \\
\midrule
\multicolumn{6}{@{}l@{}}{\hspace*{0.5em}\textbf{\textit{Panel A. Cumulative patents issued}}} \\
\midrule
Firm MQ & 0.387 & 0.681** & 0.922*** & 1.189*** & 1.453*** \\
 & (0.236) & (0.285) & (0.323) & (0.352) & (0.378) \\
Firm MQ SD & 0.774 & 1.017 & 1.366 & 1.158 & 0.548 \\
 & (1.056) & (1.325) & (1.556) & (1.793) & (1.977) \\
\midrule
Pseudo R$^2$ & 0.932 & 0.933 & 0.930 & 0.925 & 0.921 \\
Observations & 46,491 & 43,987 & 40,980 & 38,014 & 35,258 \\
\midrule
\multicolumn{6}{@{}l@{}}{\hspace*{0.5em}\textbf{\textit{Panel B. Citations per issued patent}}} \\
\midrule
Firm MQ & 0.884*** & 0.825*** & 0.751*** & 0.605*** & 0.428* \\
 & (0.217) & (0.219) & (0.220) & (0.221) & (0.225) \\
Firm MQ SD & 2.439*** & 2.540*** & 2.483*** & 2.337** & 2.262** \\
 & (0.799) & (0.874) & (0.931) & (0.995) & (1.020) \\
\midrule
R$^2$ & 0.498 & 0.482 & 0.455 & 0.425 & 0.399 \\
Observations & 18,324 & 19,372 & 19,130 & 18,429 & 17,555 \\
\midrule
\multicolumn{6}{@{}l@{}}{\hspace*{0.5em}\textbf{\textit{Panel C. Real value per issued patent}}} \\
\midrule
Firm MQ & 0.471** & 0.843*** & 1.136*** & 1.419*** & 1.692*** \\
 & (0.223) & (0.226) & (0.237) & (0.245) & (0.252) \\
Firm MQ SD & 0.667 & 1.738 & 2.261* & 2.720** & 2.685* \\
 & (0.995) & (1.114) & (1.224) & (1.329) & (1.382) \\
\midrule
R$^2$ & 0.798 & 0.777 & 0.762 & 0.751 & 0.742 \\
Observations & 18,324 & 19,372 & 19,130 & 18,429 & 17,555 \\
\midrule
Controls & Yes & Yes & Yes & Yes & Yes \\
Industry $\times$ year FE & Yes & Yes & Yes & Yes & Yes \\
\bottomrule
\end{tabular}

\caption*{\footnotesize This table reports innovation outcomes cumulated through each horizon from $t+1$ to $t+5$ with industry-by-year fixed effects. Issued-patent outcomes are cumulated from $t+1$ through the indicated horizon, in the firm-years for which that window is observable; the $t+5$ column repeats column (2) of Table~\ref{tab:innov}. The patent-count panel uses zero-inclusive Poisson pseudo-maximum likelihood; the citations-per-patent and value-per-patent panels use OLS with the definitions of Table~\ref{tab:innov}. All specifications include the controls of Table~\ref{tab:firm} and three-digit NAICS industry-by-year fixed effects. Standard errors are clustered by firm. $^{***}$, $^{**}$, $^{*}$: significance at 1, 5, and 10 percent.}
\end{table}
\clearpage

\begin{table}[htbp]
\centering
\footnotesize
\ShrinkTabulars
\caption{Innovation over Future Horizons: Firm and Industry $\times$ Year Fixed Effects}
\label{tab:innovhorfe}

\begin{tabular}{lccccc}
\toprule
 & $t+1$ & $t+2$ & $t+3$ & $t+4$ & $t+5$ \\
\midrule
\multicolumn{6}{@{}l@{}}{\hspace*{0.5em}\textbf{\textit{Panel A. Cumulative patents issued}}} \\
\midrule
Firm MQ & 0.013 & 0.433 & 0.708 & 1.042 & 1.411* \\
 & (0.482) & (0.588) & (0.684) & (0.734) & (0.742) \\
Firm MQ SD & -0.002 & -0.072 & 0.211 & -0.566 & -1.877 \\
 & (1.125) & (1.301) & (1.478) & (1.745) & (1.841) \\
\midrule
Pseudo R$^2$ & 0.948 & 0.959 & 0.964 & 0.968 & 0.972 \\
Observations & 27,242 & 25,654 & 23,896 & 22,237 & 20,545 \\
\midrule
\multicolumn{6}{@{}l@{}}{\hspace*{0.5em}\textbf{\textit{Panel B. Citations per issued patent}}} \\
\midrule
Firm MQ & 0.929*** & 0.783** & 0.545 & 0.339 & 0.077 \\
 & (0.347) & (0.368) & (0.372) & (0.374) & (0.368) \\
Firm MQ SD & -0.463 & 0.271 & 0.440 & 0.783 & 0.685 \\
 & (0.804) & (0.800) & (0.739) & (0.752) & (0.719) \\
\midrule
R$^2$ & 0.759 & 0.796 & 0.818 & 0.836 & 0.853 \\
Observations & 17,713 & 18,991 & 18,744 & 18,149 & 17,264 \\
\midrule
\multicolumn{6}{@{}l@{}}{\hspace*{0.5em}\textbf{\textit{Panel C. Real value per issued patent}}} \\
\midrule
Firm MQ & 0.140 & 0.687** & 1.052*** & 1.479*** & 1.713*** \\
 & (0.271) & (0.328) & (0.366) & (0.392) & (0.400) \\
Firm MQ SD & 0.371 & 0.419 & 0.523 & 1.384** & 1.751** \\
 & (0.552) & (0.638) & (0.653) & (0.677) & (0.681) \\
\midrule
R$^2$ & 0.948 & 0.945 & 0.948 & 0.954 & 0.961 \\
Observations & 17,713 & 18,991 & 18,744 & 18,149 & 17,264 \\
\midrule
Controls & Yes & Yes & Yes & Yes & Yes \\
Firm + industry $\times$ year FE & Yes & Yes & Yes & Yes & Yes \\
\bottomrule
\end{tabular}

\caption*{\footnotesize This table repeats Appendix Table~\ref{tab:innovhor} with firm fixed effects added. It uses the cumulative outcomes, sample, and controls of Appendix Table~\ref{tab:innovhor}, adding firm fixed effects to the industry-by-year fixed effects; the $t+5$ column repeats column (3) of Table~\ref{tab:innov}. Standard errors are clustered by firm. $^{***}$, $^{**}$, $^{*}$: significance at 1, 5, and 10 percent.}
\end{table}
\clearpage

\begin{table}[htbp]
\centering
\footnotesize
\ShrinkTabulars
\caption{Innovation with Match Quality Fixed in 2016}
\label{tab:innovhor2016}

\begin{tabular}{lccccc}
\toprule
 & $t+1$ (2017) & $t+2$ (2018) & $t+3$ (2019) & $t+4$ (2020) & $t+5$ (2021) \\
\midrule
\multicolumn{6}{@{}l@{}}{\hspace*{0.5em}\textbf{\textit{Panel A. Cumulative patents issued}}} \\
\midrule
Firm MQ & 0.353 & 0.756** & 0.831** & 0.820* & 0.772* \\
 & (0.311) & (0.348) & (0.394) & (0.430) & (0.462) \\
Firm MQ SD & -1.204 & -0.298 & 0.630 & 1.079 & 0.906 \\
 & (2.165) & (2.356) & (2.544) & (2.674) & (2.743) \\
\midrule
Pseudo R$^2$ & 0.942 & 0.942 & 0.935 & 0.925 & 0.917 \\
Observations & 2,824 & 2,824 & 2,845 & 2,845 & 2,853 \\
\midrule
\multicolumn{6}{@{}l@{}}{\hspace*{0.5em}\textbf{\textit{Panel B. Citations per issued patent}}} \\
\midrule
Firm MQ & 1.021*** & 0.998*** & 0.810*** & 0.716** & 0.724*** \\
 & (0.392) & (0.337) & (0.307) & (0.294) & (0.272) \\
Firm MQ SD & 3.794* & 3.194* & 3.183* & 3.121* & 2.467 \\
 & (2.301) & (1.937) & (1.788) & (1.725) & (1.581) \\
\midrule
R$^2$ & 0.235 & 0.231 & 0.234 & 0.229 & 0.236 \\
Observations & 1,117 & 1,247 & 1,341 & 1,389 & 1,429 \\
\midrule
\multicolumn{6}{@{}l@{}}{\hspace*{0.5em}\textbf{\textit{Panel C. Real value per issued patent}}} \\
\midrule
Firm MQ & 0.389 & 0.522 & 0.721** & 1.021*** & 0.801** \\
 & (0.348) & (0.350) & (0.342) & (0.361) & (0.365) \\
Firm MQ SD & -0.239 & 0.978 & 0.176 & 0.914 & 2.233 \\
 & (1.902) & (1.820) & (1.794) & (1.887) & (1.945) \\
\midrule
R$^2$ & 0.788 & 0.778 & 0.777 & 0.759 & 0.742 \\
Observations & 1,117 & 1,247 & 1,341 & 1,389 & 1,429 \\
\midrule
Controls & Yes & Yes & Yes & Yes & Yes \\
Industry FE & Yes & Yes & Yes & Yes & Yes \\
\bottomrule
\end{tabular}

\caption*{\footnotesize This table repeats Appendix Table~\ref{tab:innovhor} with firm match quality scored by the time-split checkpoint, which is trained on data through 2016 only. Match quality and all controls are measured in a single year, 2016, and the outcomes are cumulated from 2017 through the indicated year. Panel~A reports issued patents, estimated by zero-inclusive Poisson pseudo-maximum likelihood; Panels~B and~C report citations per issued patent and real value per issued patent, as defined in Table~\ref{tab:innov}, estimated by OLS on firms with at least one issued patent. Continuous outcomes and controls are winsorized at the first and ninety-ninth percentiles. Each specification includes the controls of Table~\ref{tab:firm} and three-digit NAICS industry fixed effects; firm and industry-by-year fixed effects cannot be estimated in a single cross-section. Standard errors are clustered by firm. $^{***}$, $^{**}$, $^{*}$: significance at 1, 5, and 10 percent.}
\end{table}
\clearpage

\begin{table}[htbp]
\centering
\footnotesize
\ShrinkTabulars
\caption{Acquisitions with Match Quality Fixed in 2016}
\label{tab:ma2016}

\begingroup
\centering
\begin{tabular}{lcccc}
   \toprule
                                & (1)             & (2)            & (3)            & (4)\\
   \midrule
   Delta MQ                     & 1.764$^{**}$    & 2.361$^{**}$   & 2.218$^{**}$   & 1.395\\
                                & (0.7760)        & (0.9214)       & (0.8839)       & (1.108)\\
   Pre-deal acquirer MQ         &                 & -1.962         & -3.183$^{*}$   & -2.322\\
                                &                 & (1.382)        & (1.775)        & (2.034)\\
   Log target relationships     &                 & 0.0096         & -0.0134        & -0.1164\\
                                &                 & (0.0832)       & (0.0876)       & (0.1214)\\
   Log acquirer assets          &                 &                & 0.0914$^{*}$   & 0.0888\\
                                &                 &                & (0.0509)       & (0.0591)\\
   Acquirer ROA                 &                 &                & -2.366$^{**}$  & -1.523\\
                                &                 &                & (1.026)        & (0.9658)\\
   Acquirer leverage            &                 &                & -1.669$^{***}$ & -1.186$^{**}$\\
                                &                 &                & (0.4963)       & (0.5352)\\
   Acquirer M/B                 &                 &                & 0.0570$^{*}$   & 0.0396\\
                                &                 &                & (0.0288)       & (0.0260)\\
   Acquirer sales growth        &                 &                & 0.1847         & 0.2328\\
                                &                 &                & (0.3296)       & (0.3367)\\
   Acquirer tangibility         &                 &                & 0.2230         & 0.5735\\
                                &                 &                & (0.6279)       & (0.8944)\\
   Acquirer R\&D/assets         &                 &                & -4.970$^{**}$  & -3.341\\
                                &                 &                & (2.078)        & (2.246)\\
   R\&D missing                 &                 &                & -0.2532        & -0.2761\\
                                &                 &                & (0.2465)       & (0.2810)\\
   Same industry                &                 &                & -0.2743$^{*}$  & -0.1085\\
                                &                 &                & (0.1511)       & (0.1906)\\
   Public target                &                 &                & -0.0011        & 0.1478\\
                                &                 &                & (0.3037)       & (0.3014)\\
   Constant                     & -0.8389$^{***}$ & -0.9646$^{**}$ & -0.9113$^{*}$  &   \\
                                & (0.1056)        & (0.3738)       & (0.4589)       &   \\
   \midrule
   Observations                 & 67              & 67             & 67             & 66\\
   R$^2$                        & 0.05948         & 0.08687        & 0.38469        & 0.51970\\
   \midrule
   Acquirer NAICS fixed effects &                 &                &                & Yes\\
   Target NAICS fixed effects   &                 &                &                & Yes\\
   \bottomrule
\end{tabular}
\par\endgroup

\caption*{\footnotesize This table repeats Table~\ref{tab:ma} with match quality scored by the time-split checkpoint, which is trained on data through 2016 only, for the deals whose pre-deal match quality is measured in 2016 and whose goodwill is recorded in 2017 (67 deals; 66 once fixed-effect singletons are dropped). The dependent variable and $\Delta\MQ$ are defined as in Table~\ref{tab:ma}. Columns (1)--(3) follow columns (1)--(3) of Table~\ref{tab:ma}; column (4) adds acquirer-industry and target-industry fixed effects (one-digit NAICS sectors). Year fixed effects do not apply to a single cross-section. Standard errors are clustered by acquirer. $^{***}$, $^{**}$, $^{*}$: significance at 1, 5, and 10 percent.}
\end{table}
\clearpage

\end{document}